\documentclass[12pt]{article}
\usepackage{hyperref}
\usepackage{array} 
\usepackage{bm}

\usepackage{epsfig}
\usepackage{graphicx}
\usepackage[CaptionAfterwards]{fltpage}
\usepackage{amsmath}
\usepackage{amssymb} % comment this when newtxmath is defined
\usepackage{amsfonts}
\usepackage{wrapfig}
\usepackage{lipsum}
\usepackage{caption}
\usepackage{tikz}
\definecolor{myred}{HTML}{FF5757}  % Crimson red

\usepackage{subcaption}
\usepackage{newtxtext,newtxmath}
\usepackage{xcolor}
\usepackage{color}
\usepackage{makecell}
\usepackage{textcomp}
\usepackage{gensymb}
\usepackage{siunitx}
\usepackage{tabularx}
\usepackage{booktabs}
\usepackage{float} % add this to preamble

\usepackage[letterpaper,margin=1in]{geometry}

\renewenvironment{abstract}
	{\quotation}
	{\endquotation}

\date{}

\makeatletter

\renewcommand{\fnum@figure}{\textbf{Fig.\,\thefigure }} % nature article web style use :
\renewcommand{\fnum@table}{\textbf{Table \upthetable}}

\makeatother

\usepackage{scicite}

\usepackage{url}

\newcommand{\mypara}[1]{\par\vspace*{1.5mm}\noindent\textbf{{#1}}}

\DeclareCaptionLabelFormat{nospace}{#1#2}

\DeclareCaptionType{extfigure}[Fig.~S]
\DeclareCaptionType{exttable}[Table S]
\def\scititle{Extreme Dynamic Symmetry Enables Omnidirectional and Multifunctional Robots}
\title{\bfseries \boldmath \scititle}

\author{
	Jiaxun Liu$^{1\dagger}$, Boxi Xia$^{1\dagger}$, Boyuan Chen$^{1, 2, 3\ast}$ \\
    \normalsize{$^{1}$Department of Mechanical Engineering and Materials Science, Duke University}\\
    \normalsize{$^{2}$Department of Electrical and Computer Engineering, Duke University}\\
    \normalsize{$^{3}$Department of Computer Science, Duke University}\\
    \normalsize{$^\ast$To whom correspondence should be addressed; E-mail: boyuan.chen@duke.edu.}\\
    \normalsize{$^\dagger$These authors contributed equally to this work.}
}

\begin{document} 

% Insert the title and author list
\maketitle

\begingroup
\renewcommand\thefootnote{}
\footnotetext[0]{\raggedright This is the author’s version of the work. It is posted here by permission of AAAS for personal use, not for redistribution. The definitive version was published in Science Robotics on May 27, 2026, DOI: https://doi.org/10.1126/scirobotics.aec1725. Published version: https://www.science.org/doi/10.1126/scirobotics.aec1725.}
\endgroup

\begin{center}
\vspace{-25pt}
\href{https://generalroboticslab.com/Argus}{Project website: generalroboticslab.com/Argus}
\end{center}

% \footnote{This is the author’s version of the work. It is posted here by permission of AAAS for personal use, not for redistribution. The definitive version was published in Science Robotics on May 27, 2026, DOI: https://doi.org/10.1126/scirobotics.aec1725. Published version: https://www.science.org/doi/10.1126/scirobotics.aec1725.}

% Abstract, in bold
% There are strict length limits, and not all formats have abstracts.
% Consult the journal instructions to authors for details.
% Do not cite any references in the abstract.
\begin{abstract} \bfseries \boldmath
Symmetry is a central organizing principle in natural systems, yet its use as a unifying design strategy in robotics has largely remained limited to geometric form. We show that symmetry can instead be leveraged at the level of dynamic actuation capability. We introduce dynamic symmetry, the uniformity of a robot’s attainable center-of-mass accelerations, and formalize it through a measure coined as dynamic isotropy. Across more than 1,000 simulated morphologies, we found that higher dynamic symmetry consistently improves trajectory tracking, task success, robustness, resiliency, and energy efficiency, with the benefits becoming most pronounced as dynamic isotropy approaches its theoretical limit. To study this regime systematically, we developed Argus, a family of spherical robots designed to explore the effects of increasing dynamic symmetry. Members of the Argus family vary in their actuation geometry and dynamic symmetry level, while sharing a common architectural principle: radially oriented linear actuators that directly shape the robot’s center-of-mass
dynamics. Among them, we build a physical 20-leg Argus variant that achieves near-extreme dynamic isotropy and demonstrates orientation-invariant locomotion, agile traversal of cluttered and deformable terrain, rapid self-stabilization, and resilience to partial actuator failures. Its distributed sensing further enables omnidirectional perception and object interaction during continuous motion. These results show that designing robots for symmetry not only in morphology but also in their attainable dynamics provides a powerful and general pathway toward agility, robustness, and multifunctionality in uncertain terrestrial and extraterrestrial environments.
\end{abstract}

\begin{FPfigure}
    \centering
    \includegraphics[width=\textwidth]{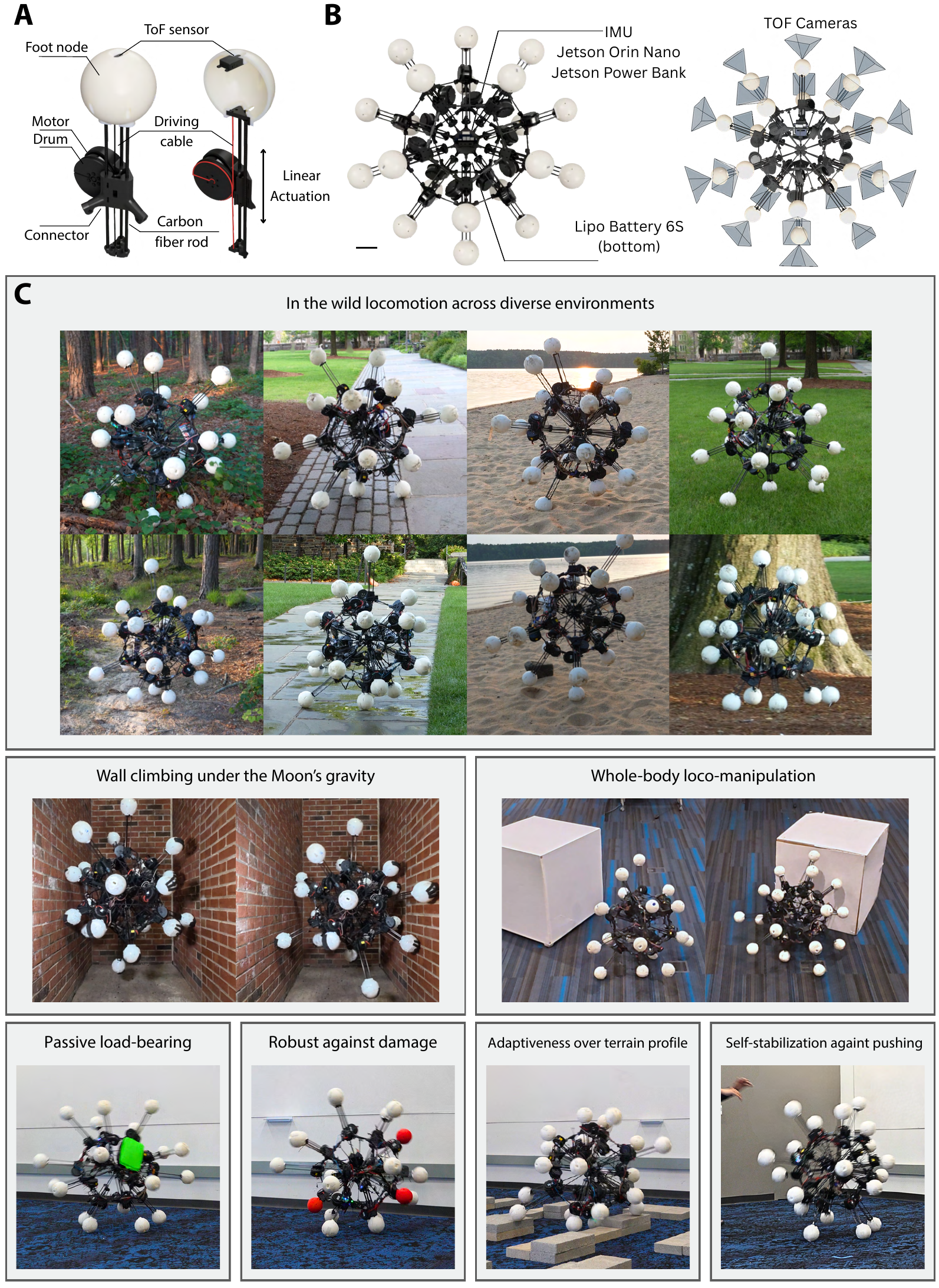} 
	\caption{{$|$ \textbf{Argus: a dynamically symmetric robot family and its 20-leg hardware realization.} \textbf{(A)}, Each leg module is identical and consists of a radially oriented 1-degree-of-freedom linear actuator driven by a cable–drum mechanism. A Time-of-Flight (ToF) depth sensor is embedded in each foot node, enabling omnidirectional perception co-located with actuation. \textbf{(B)}, The 20-leg Argus prototype arranges its actuators at the vertices of a regular dodecahedron, producing a near-spherical distribution of actuation directions and achieving near-extreme dynamic isotropy of 0.91. Computation, inertial sensing, and power modules are mounted at the center of the structure, whereas 20 outward-facing ToF cameras collectively provide omnidirectional depth measurements for perception and interaction. Scale bar, 0.1m. \textbf{(C)}, The physical robot operating across diverse indoor and outdoor environments, including grass, sand, bark, pavement, narrow corridors, and whole-body loco-manipulation settings, demonstrating that extreme dynamic symmetry enables robust omnidirectional locomotion and multifunctional behaviors in unstructured environments.}}
    \label{figure_1}    
\end{FPfigure}

\section*{Introduction}

Since the term ``robot'' was first coined over a century ago \cite{capek2004rur}, the design of robotic systems has remained a central scientific and engineering challenge. Advances in actuation, sensing, and computation have been matched by a rich diversity of mechanical designs, many of which draw inspiration from biology \cite{pfeifer2007self, trivedi2008soft,iida2016biologically,bar2003biologically,lepora2013state}. Quadrupedal robots echo the musculoskeletal architecture of dogs \cite{raibert2008bigdog,hutter2016anymal,bledt2018cheetah}; humanoid robots\cite{metta2010icub,saloutos2023design, xia2025duke,liao2025berkeley} emulate human proportions and bipedal locomotion; articulated manipulators mimic the dexterity of arms and hands \cite{mattar2013survey,ma2011dexterity}; bio-inspired underwater robots \cite{sun2022recent,baines2022multi,li2021self}, flapping-wing drones \cite{shin2024fast, chang2024bird}, and even simplified mechanisms such as parallel grippers inherit key features from their natural counterparts \cite{langowski2020soft}. The unifying theme in these approaches is imitation, aiming to reproduce the form and function of biological systems, then refining the designs to meet real-world constraints.

Here, we explore a different way of learning from nature: extracting and applying fundamental organizational principles that transcend specific species or morphologies. One such principle is symmetry, a pervasive feature of living systems across scales and taxa \cite{ocklenburg2022symmetry, enquist1994symmetry}. From the bilateral forms of vertebrates and the radial symmetry of starfish, to the helical arrangements of plant seeds and the geometric regularity of viruses, symmetry is deeply entwined with both structure and perception. In animals, control laws often exploit these symmetries to simplify coordination of whole-body actuations and gain robustness under diverse environmental conditions.

Inspired by these ideas in nature, symmetry has long played a central role in how roboticists design body configurations that move and interact with the world. Most prior work has emphasized morphological symmetry, specifically the geometric arrangement of limbs, links, or actuators. Recent studies have shown that symmetric bodies often help improve learning efficiency\cite{zhu2022sample,wang2022equivariant,wang2023robot,su2024leveraging,mittal2024symmetry}, simplify dynamics and control\cite{apraez2025morphological,yan2024learning}, or enhance maneuverability\cite{frazzoli2005maneuver}. Examples range from legged robots with bilateral symmetry, such as biped and quadruped robots, to spherical\cite{pai1994platonic, nozaki2018Continuous,gheorghe_rolling_2010,liu_design_2020} and tensegrity platforms\cite{paul2006design,vespignani2018design,kim_rolling_2020,jeong_spikebot_2024,surovik_adaptive_2021,liu2025lcrbot,chen2024learning} whose shapes exhibit higher-order geometric repetition. Despite this rich history, these approaches still remain grounded in the geometry of the robot’s body rather than in the forces and accelerations the robot can produce. 

This gap motivates a shift in perspective: from symmetry of form to symmetry of dynamic actuation capability. We refer to this idea as ``dynamic symmetry'' -- the property that a robot can generate forces and accelerations with uniform magnitude in all directions. 
Where morphological symmetry ensures that no spatial orientation is privileged, dynamic symmetry ensures that no direction of action is privileged. 
This can enable a fundamentally different suite of behaviors: robust omnidirectional locomotion, rapid disturbance rejection from arbitrary angles, resilient recovery from broken actuators or sensors, seamless transition between tasks, and natural multifunctionality in complex environments.

To quantify dynamic symmetry, we introduced a theoretical construct for rigid-body legged robots that we call dynamic isotropy. Classical isotropy theory in robotics, such as manipulability-based isotropy and Jacobian condition number analyses \cite{salisbury1982articulated,yoshikawa1985manipulability,klein1987dexterity,ma1990concept,klein1991spatial,kim2007systematic}, is purely kinematic and applies only to end-effector velocity mappings of serial manipulators. These metrics characterize how uniformly joint motions map to Cartesian velocities, but they do not incorporate actuator force limits, mass properties, thrust directions, or the geometry of whole-body dynamic actuation capability. Although previous dynamic isotropy\cite{ma1990concept,ma1993optimum} and dynamic manipulability ellipsoids \cite{yoshikawa1985dynamic} introduce inertia, they remain local and end-effector–centric, capturing neither full-body acceleration feasibility nor global actuation symmetry.

In contrast, we define dynamic isotropy directly on the robot’s attainable center-of-mass(CoM) acceleration set. It measures how isotropic the feasible acceleration space becomes as the robot’s actuation directions increase in number, uniformity, and spatial distribution. This yields a global characterization of whole-body dynamics, which reflects how evenly the robot can act on the environment, rather than being limited to how its joint motions map to velocity. Dynamic isotropy thus provides a principled way to study and design dynamic symmetry as an axis of robot capability.

In this article, we investigate extreme dynamic symmetry as a design principle for omnidirectional and multifunctional rigid-legged robots (Movie 1). Our central hypothesis is that, as dynamic isotropy approaches its theoretical limit, a robot gains enhanced dynamic actuation capabilities, enabling it to become an omnidirectional and multifunctional machine. We test this hypothesis through both theoretical and empirical studies. First, we derive and analyze dynamic isotropy as a performance measure grounded in whole-body dynamics and control. We then introduce Argus as the experimental platform, a class of spherical robot designs(Fig. S\ref{fig:dof_vs_energy_isotropy}) that feature 1-degree-of-freedom linear actuated legs(Fig.~\ref{figure_1}A) oriented toward the robot's center and directly alter its center of mass(CoM) dynamics. With Argus, we conduct large-scale simulation sweeps across thousands of morphologies to investigate how increasing dynamic symmetry alters the attainable acceleration set, enhances robustness, resiliency, and task performance. Finally, we realize a physical Argus variant with 20 legs and 20 cameras at its foot nodes(Fig.~\ref{figure_1}B) with near-extreme dynamic symmetry and demonstrate capabilities that exceed those of prior spherical and tensegrity platforms across locomotion, robustness, agility, resilience, and multifunctionality(Fig.~\ref{figure_1}C). Beyond locomotion, we show that extreme dynamic symmetry pairs naturally with omnidirectional perception(Fig.~\ref{figure_1}C): the robot can maintain omnidirectional perception under aggressive maneuvers. Together, dynamic isotropy and omnidirectional sensing produce a versatile, multifunctional system capable of seamless transitions between locomotion, reorientation, environment interaction, and perception-driven tasks such as whole-body loco-manipulation. Overall, our work establishes dynamic isotropy as a theoretical framework for whole-body robotic capability and demonstrates that pushing dynamic symmetry toward its limit unlocks versatile behaviors in mobility, robustness, perception, and multifunctionality.

\begin{FPfigure}
    \centering
    \includegraphics[width=1\textwidth]{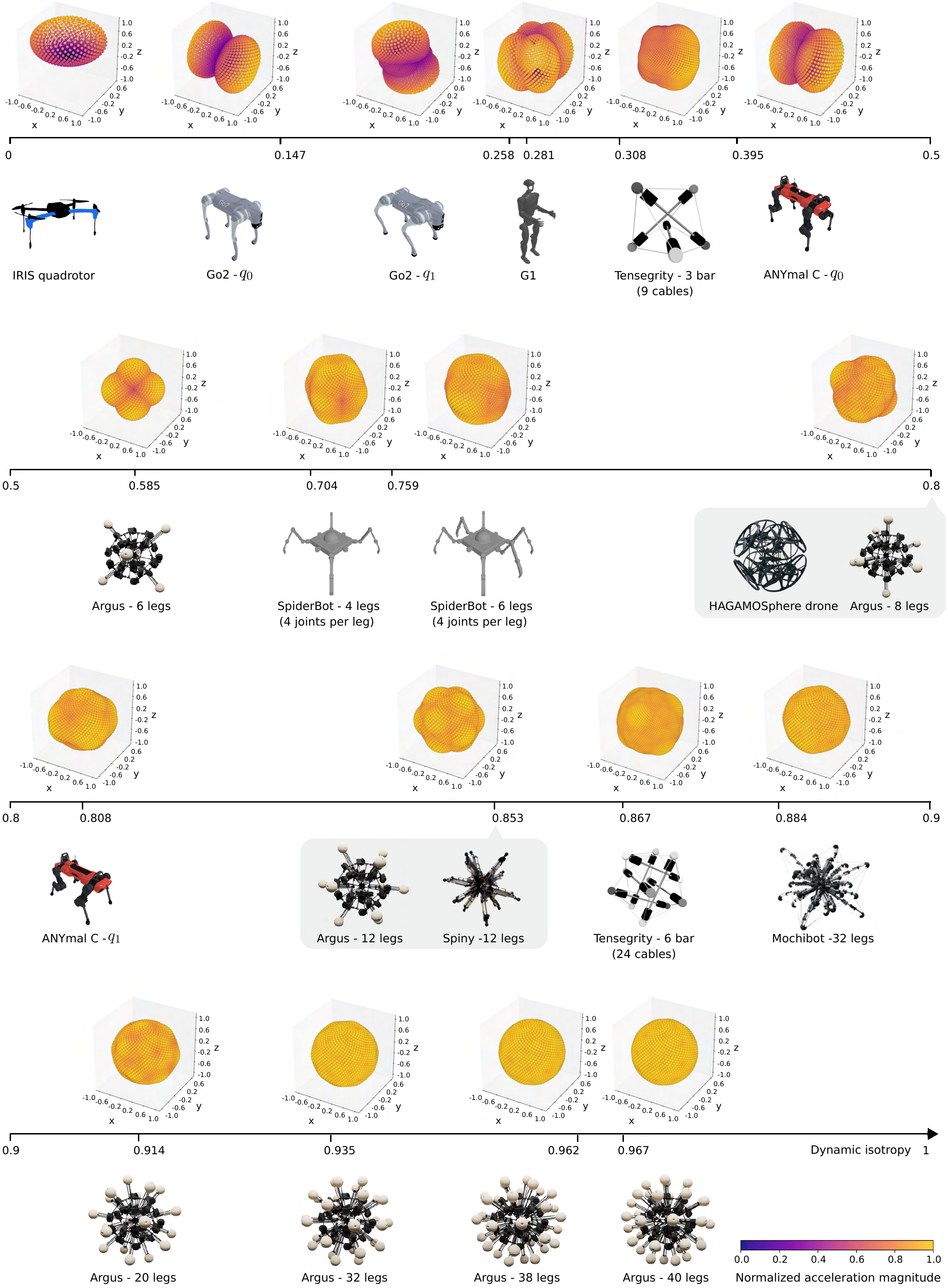} 
    \caption{{$|$ \textbf{Dynamic isotropy across different robots and joint configurations.} These robots include IRIS quadrotor, Unitree Go2, Unitree G1, tensegrity robots\cite{chen2024learning}, ANYmal C, SpiderBots\cite{dasgupta_spiderbot_deeprl}, HAGAMOSphere (research prototype, DIC Corporation), Spiny\cite{nozaki2017Shape}, and Mochibot\cite{nozaki2018Continuous}. For each robot, the 3D scatter plot shows attainable CoM acceleration magnitudes sampled over 2,048 uniformly distributed directions on the unit sphere.  
    Each point lies along a direction $\mathbf{u}$, with radial distance and color indicating the normalized maximum acceleration magnitude $a_{\max}(\mathbf{u})$ relative to the global maximum. Systems with low dynamic isotropy exhibit highly lopsided, streaked acceleration clouds, whereas systems with high dynamic isotropy display nearly spherical, uniformly bright acceleration clouds. Most of the robot designs exhibit dynamic isotropy lower than 0.9. In contrast, Argus design can achieve a higher dynamic isotropy score, approximating the maximum value of 1 in the last row. }}
    \label{fig:isotropy_robots}
\end{FPfigure}

\section*{Results}

\subsection*{Dynamic symmetry and dynamic isotropy}

Robots differ not only in appearance but also in how they can interactwith the environment in terms of their dynamic actuation capabilities. We define dynamic symmetry as the uniformity of a robot’s attainable CoM accelerations: a dynamically symmetric robot can accelerate itself with uniform authority in many, ideally all, directions.

Formally, we model the relationship between joint torques $\boldsymbol{\uptau}$ and CoM accelerations $\mathbf{a}_c$ as

\begin{equation}
\mathbf{a}_c = \mathbf{A}(\mathbf{q}) \, \boldsymbol{\uptau},
\end{equation}

where $\mathbf{A}(\mathbf{q})$ is a configuration-dependent matrix that incorporates actuation directions, actuator limits, and the robot’s mass distribution (see the "Dynamic Symmetry and Dynamic Isotropy" and "Theoretical Analysis of Dynamic Isotropy for Stability, Robustness, and Control Efficiency" sections for the full derivation). For a unit direction $\mathbf{u} \in \mathbb{S}^2$, we define $a_{\max}(\mathbf{u})$ as the maximum CoM acceleration attainable along $\mathbf{u}$ under actuator constraints. Sampling many directions on the unit sphere yields a discrete approximation of the robot’s attainable acceleration set.

We introduce ``dynamic isotropy'' $\upeta$ as

\begin{equation}
\upeta = \frac{a_{\min}}{a_{\max}},
\end{equation}

where $a_{\min}$ and $a_{\max}$ are the minimum and maximum values of $a_{\max}(\mathbf{u})$ across all sampled directions.  
When $\upeta \rightarrow 1$, the robot can accelerate almost equally well in every direction; when $\upeta$ is small, certain directions are much harder to actuate than others.

Fig.~\ref{fig:isotropy_robots} illustrates dynamic isotropy among different robots and joint configurations. We present the detailed description and analysis of the dynamic isotropy of different robots in the "Supplementary Methods" section of the Supplementary Materials. We also present a more detailed analysis of the robot’s dynamic isotropy under different configurations during forward motion in Fig.S\ref{fig:isotropy_trajectory}. These examples also demonstrate that dynamic symmetry is a distinct property that is not guaranteed by geometric symmetry alone. For each system, we visualize the distribution of attainable acceleration magnitudes over 2,048 uniformly sampled directions. Robots with low dynamic isotropy exhibit lopsided, streaked acceleration clouds with large variations in magnitude, indicating pronounced directional preferences. In contrast, robots with higher dynamic isotropy show nearly spherical, uniformly colored clouds, reflecting an almost direction-agnostic ability to accelerate the CoM.

It is worth noting that certain scores may appear unintuitive, such as the low dynamic isotropy of conventional quadrotor drones. This outcome arises because dynamic isotropy quantifies the uniformity of instantaneously attainable linear accelerations of the CoM, which is different than overall maneuverability. Although quadrotors can execute agile three-dimensional motions through attitude reorientation, their coplanar rotor configuration restricts certain directions of instantaneous thrust generation. As a result, lateral acceleration authority is limited without prior reorientation, leading to reduced dynamic isotropy. As seen in our plot, designs such as spherical drones achieve higher dynamic isotropy than conventional quadrotors.

Overall, these results demonstrate that the uniformity of instantaneous attainable CoM accelerations can be consistently characterized across diverse robots, including aerial robots, tensegrity systems, legged robots, and humanoids, using dynamic isotropy as a unified theoretical framework. Importantly, as discussed in the next section, we further focus on understanding the benefits for legged robotic systems operating near extreme dynamic isotropy.

Our visualizations highlight that dynamic symmetry is a property of force and acceleration capability, distinct from geometric symmetry alone. Moreover, our theoretical analysis shows that high dynamic isotropy ensures that the robot’s acceleration map is well conditioned with orientation-invariant stability margins, uniform robustness to external disturbances, and balanced control efforts (see ``Materials and Methods - Theoretical Analysis of Dynamic Isotropy for Stability, Robustness, and Control Efficiency'').

\subsection*{Argus: a family of robot designs to study dynamic symmetry}

To systematically explore extreme dynamic symmetry, we develop a family of spherical robots named Argus. Each Argus variant consists of 1-degree-of-freedom linear legs mounted on a spherical frame and oriented radially toward the robot’s CoM. Because each leg can only apply force along its own axis, the spatial distribution of leg directions directly shapes the attainable acceleration set: adding more legs increases actuation density, and distributing them more uniformly over the sphere increases dynamic isotropy.

We generate Argus morphologies by varying both the number of legs and their placement. Two complementary construction strategies span a wide range of dynamic isotropy: Thomson energy \cite{glasser1992energies} optimization, which distributes legs on a sphere by minimizing the sum of inverse pairwise distances, producing near-uniform geometric layouts; and random sampling over a sphere to explore asymmetric morphologies. Across these designs, dynamic isotropy values range from 0.32 to 0.97 (Fig. S\ref{fig:dof_vs_energy_isotropy}), enabling us to systematically study dynamic symmetry from low to near-extreme regimes. 

\begin{figure}[!t]
    \centering
    \includegraphics[width=0.94\textwidth]{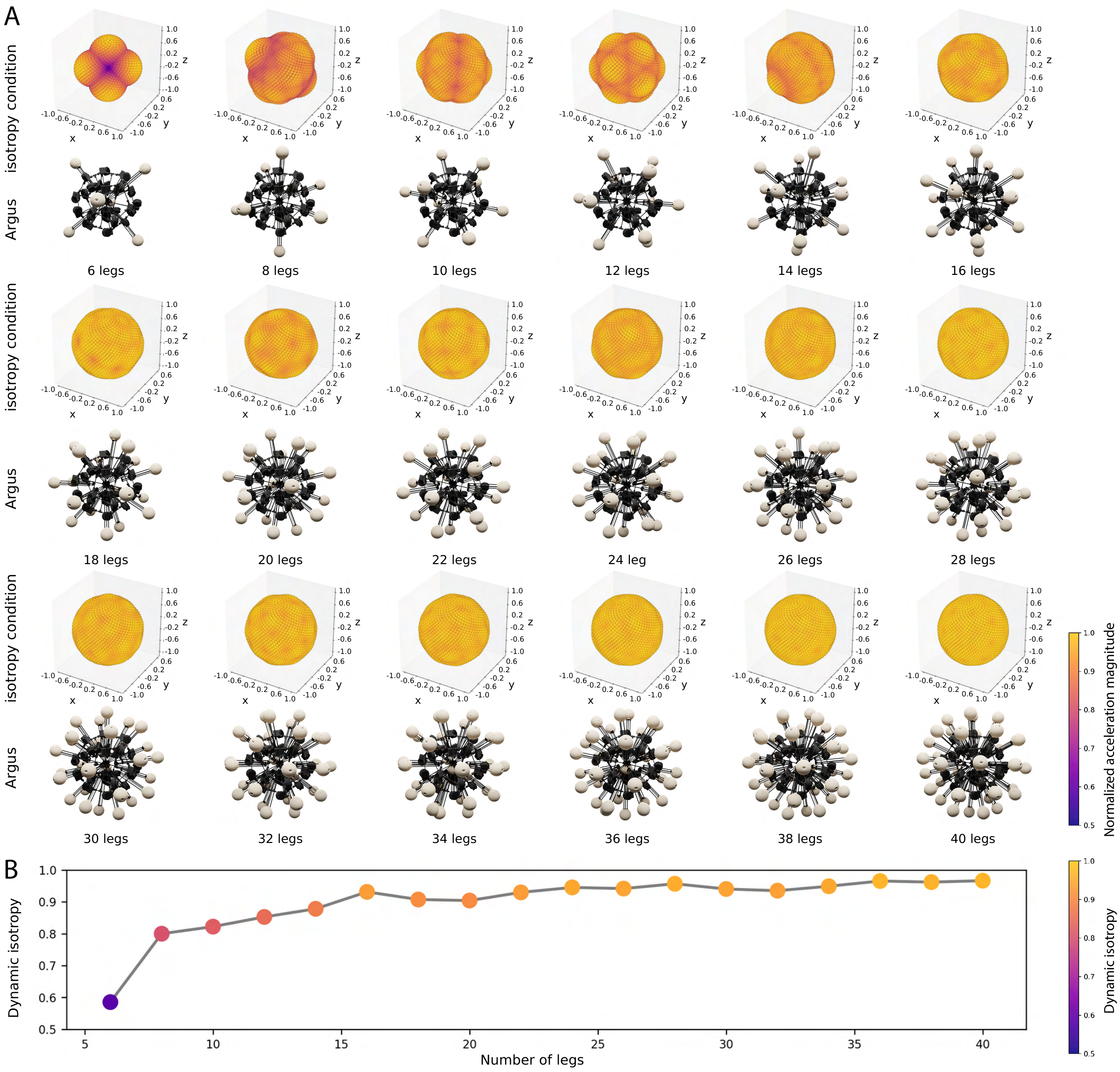}
    \caption{{$|$ \textbf{Dynamic isotropy in symmetric Argus variants with different numbers of legs.} \textbf{(A)}, Visualization of attainable CoM acceleration magnitudes for symmetric Argus variants with increasing numbers of radially oriented legs. Each cloud is constructed from 2,048 uniformly sampled directions; point radius and color indicate normalized maximum acceleration magnitude along that direction. As leg count increases, the attainable acceleration clouds becomes more uniform and spherical. \textbf{(B)}, Dynamic isotropy score $\upeta$ for the corresponding variants. Increasing the number of legs monotonically raises $\upeta$ but with diminishing returns: isotropy begins to plateau once the number of legs exceeds roughly 16 to 22, indicating that these designs are already close to the theoretical limit of uniform CoM acceleration for this actuation model.}}
    \label{fig:isotropy_different_legs}
\end{figure}

Figure~\ref{fig:isotropy_different_legs} compares symmetric Argus variants with different numbers of legs. As the leg count increases from 6 to 40, the attainable acceleration clouds become progressively more spherical and uniform, and the dynamic isotropy score $\upeta$ rises accordingly. However, this increase exhibits a plateau: beyond approximately 16 to 22 legs, additional actuators yield only marginal gains in dynamic isotropy. This trend suggests a region of diminishing returns, where the robot is already close to the theoretical isotropy limit imposed by its actuation model. Although configurations with more legs have higher redundancy, their dynamic isotropy remains relatively similar, suggesting that dynamic isotropy captures effects not reflected by redundancy alone and provides an additional criterion for selecting an appropriate leg count to achieve strong performance.

For the physical prototype, manufacturability and space for electronics impose additional constraints. Polyhedral shapes with near-extreme dynamic symmetry are ideal for physical robot construction since the edges can provide stiff support for each actuator and the faces can house electronics and wiring. We therefore built a 20-leg Argus robot by finding the closest dodecahedral configuration with near extreme dynamic isotropy score of 0.91. The design is modular by mounting identical cable-driven linear actuators at the 20 vertices of a dodecahedron (see ``Materials and Methods – Mechanical Design'').

\begin{figure}[t!]
    \centering
    \includegraphics[width=1\textwidth]{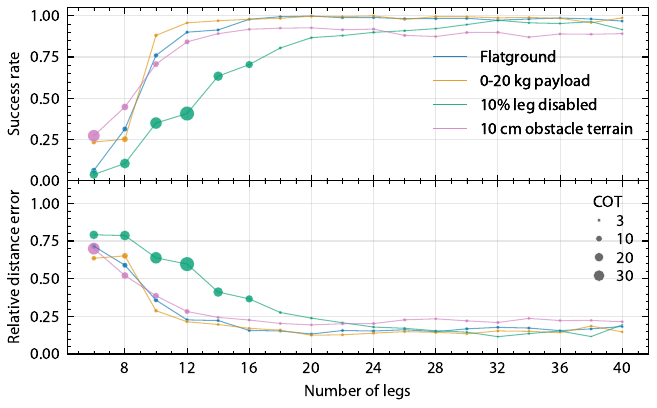}
    \vspace{-20pt}
    \caption{{$|$ \textbf{Performance scaling with dynamic isotropy in symmetric Argus variants.} For 18 symmetric Argus variants with 6 to 40 legs, we plot trajectory tracking error and COT across all four tasks. As dynamic isotropy rises, both tracking error and COT decrease, whereas task success rates increase. These improvements saturate between roughly 12 and 22 legs, closely aligning with the regime where dynamic isotropy itself begins to plateau (around 16–22 legs).}}
    \label{fig:argus_min_energy}
\end{figure}

\subsection*{Dynamic symmetry vs. performance in large-scale simulations}

Our key hypothesis is that robots with higher dynamic symmetry should perform better across diverse tasks, especially as isotropy approaches its theoretical limit. To test this, we conducted large-scale robot learning experiments in simulation, evaluating Argus variants in locomotion and interaction tasks under varying dynamic isotropy.

We consider four representative tasks: velocity tracking on flat ground; locomotion with $10\%$ of legs disabled; locomotion while carrying payloads between 0 and \SI{20}{\kilogram}; and traversal over discrete obstacle terrain with obstacles up to \SI{10}{\centi\meter} high. For each morphology, we trained a locomotion policy using deep reinforcement learning and evaluated trajectory tracking error, task success rate, and cost of transport (COT) over 5-second trajectories at a commanded speed of $v_{\text{cmd}} = 0.8~\text{m/s}$ (see ``Materials and Methods – Policy Training, Tasks, and Evaluations'').

We first analyzed 18 symmetric Argus variants with leg counts between 6 and 40, each constructed via Thomson energy minimization to yield highly uniform geometric layouts. As shown in Fig.~\ref{fig:argus_min_energy}, performance improved with the number of legs across all four tasks: tracking errors decreased, success rates increased, and COT dropped. Notably, the onset and plateau of these performance gains closely track the corresponding rise and saturation of dynamic isotropy in Fig.~\ref{fig:isotropy_different_legs}. Beyond around 16 to 22 legs, both isotropy and task performance show diminishing returns, indicating that pushing isotropy toward its limit offers substantial benefits up to the point where the attainable acceleration field is already nearly isotropic.

Spatial redundancy has been studied as a key mechanism for reliable legged robot locomotion. To study the relationship between our dynamic symmetry with spatial redundancy (in our case, the number of legs of Argus variants), we next generated 512 morphological variants for each of the 12-, 20-, and 32-leg configurations,  for a total of 1,536 Argus variants. Leg directions were sampled to span a broad range of dynamic isotropy values between 0.25 and 0.93 while keeping the number of legs fixed.  Representative configurations are visualized in Fig. S\ref{fig:dof_vs_energy_isotropy}.

\begin{figure}[t!]
    \centering
    \includegraphics[width=1\textwidth]{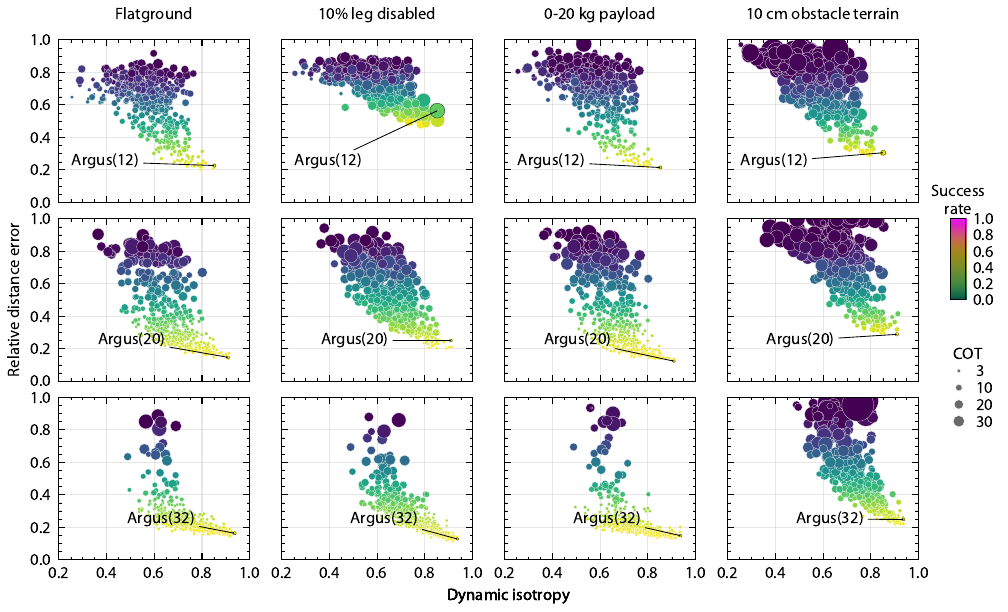}
    \vspace{-20pt}
    \caption{{$|$ \textbf{Effect of dynamic isotropy on performance across randomized Argus morphologies.} We evaluate 512 morphological variants each for 12-, 20-, and 32-leg Argus configurations (1,536 total robots), spanning dynamic isotropy scores between 0.25 and 0.93. Each point corresponds to one morphology, with the x-axis showing dynamic isotropy $\upeta$ and the y-axis showing trajectory tracking error, task success rate, and COT, for all four tasks. Morphologies with higher dynamic isotropy consistently achieve lower tracking error, higher success rates, and reduced COT for all leg counts. Labeled ``Argus'' points indicate the fully symmetric designs, which lie near the Pareto front of the dynamic symmetry–performance trade-off, highlighting extreme dynamic symmetry as a powerful design axis for robust and efficient locomotion.}}
    \label{fig:argus_sim_isotropy}
\end{figure}

Fig.~\ref{fig:argus_sim_isotropy} shows the resulting performance as a function of dynamic isotropy. Across all three leg counts and all four tasks, higher isotropy consistently correlates with lower tracking errors, higher success rates, and reduced COT. The fully symmetric Argus designs (labeled points) lie near the Pareto frontier of this dynamic isotropy–performance space, confirming that extreme dynamic symmetry brings functionally advantageous.

We further analyzed the role of redundancy by comparing performance trends across leg counts (Fig. S\ref{fig:isotropy_redundancy}). In flat-ground and payload-carrying tasks, additional legs provided benefits in the mid-range of isotropy ($0.5 \lesssim \upeta \lesssim 0.9$), but their marginal advantage diminished as isotropy approaches its extreme. Redundancy remained beneficial for leg-disabling scenarios, where extra legs provided alternative contact configurations that allowed locomotion to be preserved despite hardware failures. On discrete terrain, the number of legs played a smaller role, whereas dynamic isotropy continues to be a strong predictor of success, reflecting the importance of uniform acceleration capability in environments that demand frequent reconfiguration and contact changes.

\subsection*{Agile omnidirectional locomotion over complex terrains in the physical Argus}

We trained the locomotion policy using deep reinforcement learning in a large-scale distributed simulation (see `Policy Training, Tasks and Evaluations' in Materials and Methods). In randomized velocity tracking tasks with commanded speeds from \SI{-0.8}{\meter/\second} to \SI{0.8}{\meter/\second} along both x and y axes, Argus achieved an average tracking error of \SI{0.3}{\meter/\second}, with a maximum error of \SI{0.35}{\meter/\second} at the highest tested rolling speed of \SI{1.13}{\meter/\second}. When deployed on the physical robot directly, the same policy followed user-commanded velocities and trajectories, including sharp \SI{90}{\degree} turns (Movie S1) and arbitrary path tracking (Fig.~\ref{figure_2}A, Movie S1). This agility arises from real-time redistribution of contact forces across isotropic leg configurations, which counteract momentum and maintain balance without requiring explicit body reorientation for alignment. Despite being trained only on flat terrain, the policy transferred zero-shot to unstructured environments, including concrete bricks, grass, dense foliage with high vegetation, soft sand, and slippery wet surfaces (Fig.~\ref{figure_1}C, Movies S6-S7). Stability was preserved under orientation disturbances from ground obstacles (Fig. S\ref{fig:extend_invariant_orientation}, Movie S7), a result of its morphology's invariance to rotation.

\begin{FPfigure}
    \centering
    \includegraphics[width=\textwidth]{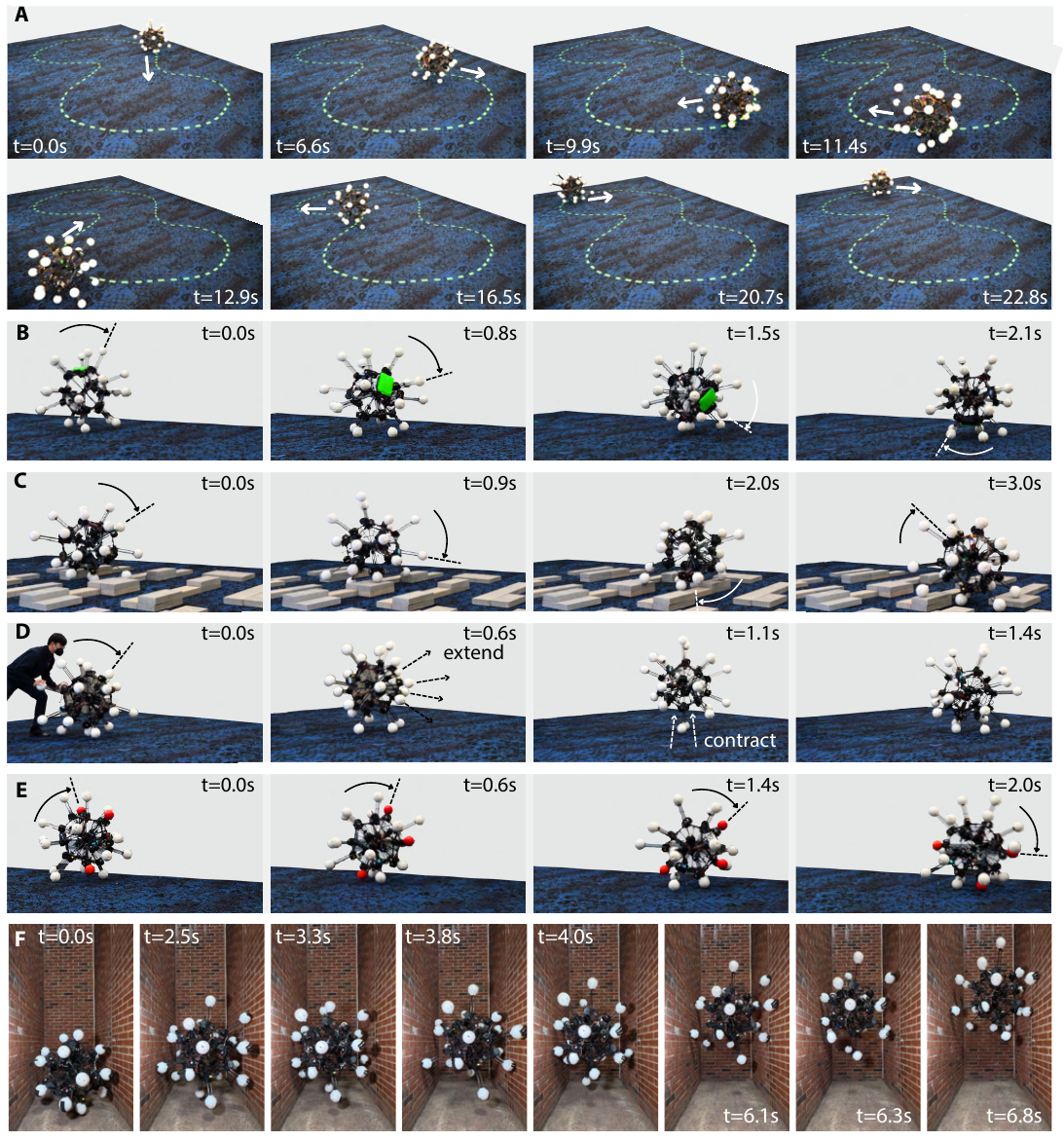} 
    \caption{{$|$ \textbf{Agile, robust, and resilient locomotion of the 20-leg physical Argus.} \textbf{(A)}, Argus tracks arbitrary user-commanded trajectories while maintaining continuous rolling motion, exploiting its nearly isotropic actuation to steer without explicit body reorientation. \textbf{(B)}, Argus transports a \SI{4.5}{\kilogram} (10 lbs) payload mounted asymmetrically on a single face while preserving most of its commanded forward speed; the dotted line traces a representative foot node (also in C–E). \textbf{(C)}, Argus traverses discrete terrain by dynamically reconfiguring which legs make contact, morphing its shape to maintain rolling continuity over obstacles. \textbf{(D)}, Under lateral pushes, Argus extends legs opposite to the disturbance to generate stabilizing torques and remain in place. \textbf{(E)}, Argus maintains locomotion despite one to three disabled legs by reorienting its body to bring healthy legs into ground contact, demonstrating resilience to partial hardware failure. \textbf{(F)}, In a lunar-gravity environment, Argus climbs between parallel walls by alternating bracing and thrusting motions using different subsets of legs, leveraging whole-body isotropic actuation to achieve vertical motion in confined spaces.}}
    \label{figure_2} 
\end{FPfigure}

When trained to traverse discrete terrain, Argus exploited its dense leg array to morph its shape, conform to the terrain's profile, and generate propulsion to track the commanded velocity. It achieved a $70\%$ success rate over discrete obstacles with \SI{0.1}{\meter} maximum height while tracking \SI{0.8}{\meter/\second} velocity commands (Fig. S\ref{fig:argus_12_20_32_sim_eval}D) in simulation, and a \SI{83.3}{\%} success rate with \SI{0.12}{\meter} maximum height while tracking \SI{0.6}{\meter/\second} velocity commands (15 out of N=18 trials) in physical experiments(Fig.~\ref{figure_2}C, Movie S5). In cases where Argus struggled due to high obstacles, it learned a recovery strategy by first adopting a stable retracted posture and then rapidly extending its legs to generate recovery momentum to push its body over the obstacle (Fig. S\ref{fig:extend_discrete_terrain}). Such adaptive behavior was facilitated by its ability to uniformly engage with the terrain, redistribute contact forces through redundant body configurations, and execute stabilization policies without requiring a specific body alignment.

Near-extreme dynamic isotropy in the design of Argus also enables passive load-bearing, self-stabilization, and fault tolerance capabilities. Its spherical structure supports the payload passively, allowing the robot to maintain stable motions with relative changes to its actuation strategy. In simulation, Argus transported a \SI{40}{\kilogram} payload, which is near two times of its own body mass (\SI{23.4}{\kilogram}), while retaining $68.4\%$ of its nominal performance without payload, at a commanded rolling speed of \SI{0.8}{\meter/\second} (Fig. S\ref{fig:argus_12_20_32_sim_eval}C). When it carried \SI{20}{\kilogram} payloads, $85.7\%$ performance was retained (Fig. S\ref{fig:argus_12_20_32_sim_eval}C). In physical testing, a payload was mounted asymmetrically on a single face of the carbon-fiber rod frame, which created a much more challenging and structurally demanding condition than the balanced simulation setup. Under this configuration, Argus carried \SI{4.5}{\kilogram} (10 lbs), maintaining \SI{96.3}{\%} of the commanded velocity of \SI{0.6}{\meter/\second} (Fig.~\ref{figure_2}B, Movie S4).

Argus achieves self-stabilization by dynamically shifting its center of mass and actively interacting with the environment. Legs can be synchronized to approximate a sphere for rolling motions, but they can also extend asymmetrically to resist motion. When dropped at \SI{2}{\meter/\second}, Argus stabilized with \SI{1.5}{\meter} (Fig. S\ref{fig:argus_12_20_32_sim_eval}A) displacement (\SI{109.5}{\%} of its maximum body length). In contrast, the purely spherical structure without coordinated leg control required \SI{2.13}{\meter} to stabilize which is \SI{42.0}{\%} worse than the performance with control policy. Under external pushing (Fig.~\ref{figure_2}D), Argus quickly responded by extending its legs in the opposite direction of the push to remain in place (Movie S3). These results highlight the role of the near-extreme isotropic leg distribution in maintaining robustness and stability under dynamic perturbations.

Resiliency against actuator failure emerged from both the redundancy and isotropy of Argus {(Fig. S\ref{fig:isotropy_redundancy}). When certain actuators failed, Argus could redistribute actuation loads uniformly to preserve functionality despite localized failures (Fig. S\ref{fig:extended_disabled_legs}). Argus retained $95\%$ and $85\%$ of original velocity at \SI{0.8}{\meter/\second} in simulation with two ($10\%$) and four ($20\%$) legs disabled respectively (Fig. S\ref{fig:argus_12_20_32_sim_eval}B). In physical experiments, forward rolling continued with one to three disabled legs (Fig.~\ref{figure_2}E). If the disabled configuration reduced efficiency, Argus reoriented its body to bring functional legs into ground contact, thereby restoring propulsion without requiring directional preference (Movie S2).

The same isotropic actuation principle enables climbing between parallel walls under the Moon's gravity (Fig.~\ref{figure_2}F, Fig. S\ref{fig:extended_wall_climbing}). Argus can extend multiple of its opposite legs to brace its body between the walls while using legs positioned diagonally to the wall surfaces to push upward. By alternating between bracing and pushing motions, Argus learned to generate upward thrust and climb effectively. In simulation, Argus climbed at an average of \SI{0.258}{\meter/\second} under the Moon's gravity at $85.5\%$ success rate, and achieved an average \SI{0.308}{\meter/\second} climbing velocity among successful trials. Our physical experiments also demonstrated that Argus could climb parallel walls with an average speed of \SI{0.238}{\meter/\second} under the Moon's gravity (Movie S8), suggesting its potential for low-gravity mobility in space exploration. These results illustrate how near-extreme dynamic symmetry supports agile locomotion not only on horizontal terrain but also in constrained, low-gravity environments relevant to space exploration.

\subsection*{Omnidirectional environmental awareness and whole-body loco-manipulation in the physical Argus}

Beyond locomotion and whole-body dynamics, we further augment the physical Argus with rich environmental awareness. Due to its modular and symmetric design, each of the 20 foot nodes can accommodate a compact time-of-flight (ToF) depth camera. Equipping all feet with identical sensors yields a radially distributed sensing array that provides nearly uniform coverage of the surrounding space. By transforming the depth information from the 20 ToF sensors (see `Mechanical design - Sensors' in Materials and Methods) into the world coordinate frame, the resulting point cloud forms a consistent 3D representation of the environment that accurately captures the surroundings (Fig.~\ref{figure_4}A,B), regardless of the robot's orientation.

\begin{figure}[t!]
  \centering
  \includegraphics[width=\textwidth]{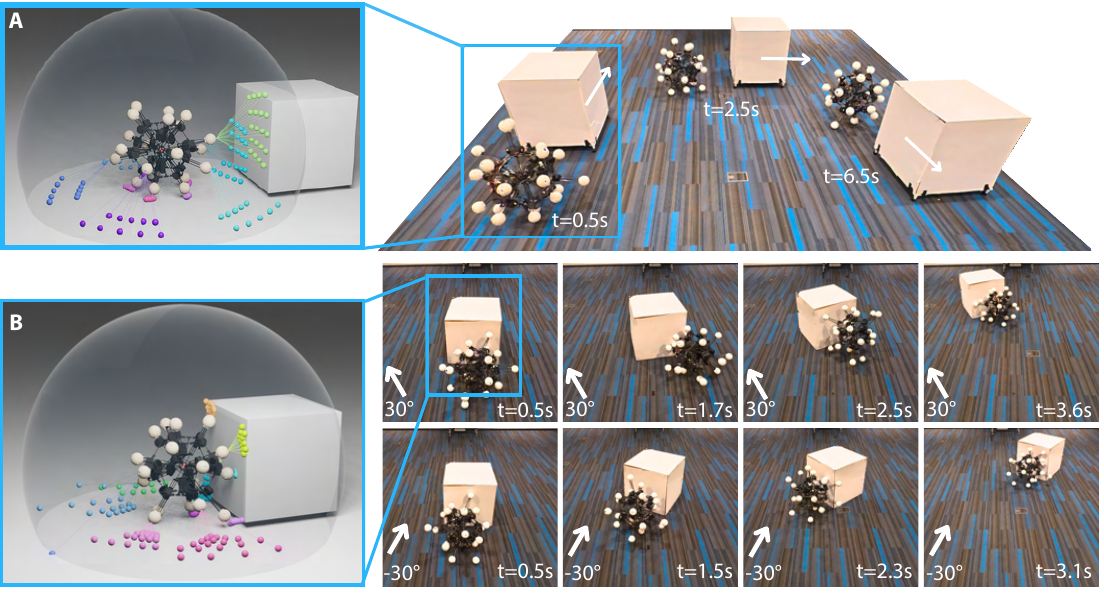}
  \caption{$|$ \textbf{Omnidirectional sensing enables object tracking and whole-body loco-manipulation.} 
  \textbf{(A)}, Argus tracks a \SI{1}{\meter} cube under varying velocity commands using depth measurements from 20 foot-mounted ToF cameras fused into a global point cloud.
  \textbf{(B)}, Using the estimated object state, Argus simultaneously pushes and follows the cube by coordinating whole-body actuation and continuous perception.}
  \label{figure_4}
\end{figure}

Prior platforms that exploit approximate spherical shapes for multi-directional motion include tensegrity robots \cite{paul2006design,vespignani2018design,kim_rolling_2020,jeong_spikebot_2024,surovik_adaptive_2021,liu2025lcrbot} and certain spherical robots \cite{pai1994platonic,nozaki2018Continuous,gheorghe_rolling_2010,liu_design_2020}. Although these systems have demonstrated intriguing capabilities, they remain fairly limited in practice: many rely on hand-crafted or non-adaptive control policies \cite{zheng2021robustness,paul2006design}, require off-board computation or power \cite{paul2006design,kim_rolling_2020}, produce relatively low forces due to small actuators \cite{nozaki2017Shape,jeong_spikebot_2024}, operate in restricted environments or task sets \cite{surovik_adaptive_2021,zheng2021robustness}, and often lack external perception \cite{xu2024physics,yang_general_2023} that could fully leverage their morphology. In contrast, our physical Argus leverages our theoretical finding of near-extreme dynamic symmetry with high-thrust onboard actuation and fully onboard distributed sensing, allowing it to exploit its uniform dynamic acuation capabilities for agile, perception-guided interaction with the environment.

When tasked with tracking a target object, Argus continuously estimated the object's velocity from the point cloud observation and achieved a success rate of \SI{89.9}{\%} and a mean tracking error of \SI{0.167 \pm 0.120}{\meter/\second} in simulation (Fig. S\ref{fig:extend_object_tracking_eval}A). In physical experiments, Argus achieved a success rate of \SI{36.8}{\%} (14 out of N=38 trials), detecting and tracking a moving target of \SI{1}{\meter} cube in various directions under the same initial pose without adjusting orientation (Movie S9). Note that the drop in real-world success rate is primarily due to sensing limitations rather than control failure. Most unsuccessful trials stem from thermal degradation in the ToF modules, which causes response delays and desynchronization across the 20 cameras during repeated experiments. These failures typically appear at the very beginning of a trial due to sensor heating from prior experiments, whereas runs that begin with stable, synchronized sensing almost always successfully complete continuous tracking. We acknowledge this as a limitation of existing sensors and highlight it as an important consideration for future sensor integration and robot design choices for deployment.

When whole-body dynamic symmetry is combined with uniformly distributed perception, Argus extends its functionality to more complex interactions, such as whole-body loco-manipulation. In an object-pushing task, Argus exploited lateral legs to apply directional forces while reshaping ground-contacting legs to stabilize and guide the object. In simulation, Argus achieved a \SI{91.5}{\%} success rate with a mean direction error of \SI{0.256 \pm 0.207}{\radian} over ten seconds of continuous pushing (Fig. S\ref{fig:extend_object_tracking_eval}B). The learned policy was successfully transferred to the real world (Fig.~\ref{figure_4}B, Movie S10) with a \SI{39.4}{\%} success rate (13 out of N=33 trials). Because actuation capability is nearly uniform in all directions, Argus maintains multiple contact points with both the ground and the object while preserving continuous perception, enabling seamless transitions between locomotion and manipulation. Similarly, most real-world failures stemmed from overheating-induced delays in ToF cameras rather than policy breakdown, suggesting that more thermally robust sensing hardware will substantially improve real-world success rates.

\section*{Discussion}

We introduced dynamic symmetry as a possible unifying principle for designing robots whose actuation capabilities remain uniform across all directions, and we formalized this idea through a whole-body measure that we call dynamic isotropy. Across more than 1,500 simulated morphologies and extensive physical experiments with a 20-leg prototype, we showed that high dynamic isotropy, especially when it's close to its theoretical extreme, enables accurate velocity tracking, robust locomotion on diverse terrain, self-stabilization under perturbations, load carrying, resilience to actuator failures, and adaptive whole-body manipulation.  Argus, our family of spherical robots built from simple, identical, radially oriented linear actuators, demonstrates that pushing dynamic symmetry toward its theoretical limit makes robots gain enhanced dynamic actuation capabilities, enabling robots to become omnidirectional and multifunctional machines.

Our results reveal that dynamic symmetry serves as a powerful organizing principle across multiple dimensions of robotic functionality. Mechanically, increasing dynamic isotropy produces more uniform whole-body interaction forces, enabling orientation-invariant locomotion and stable behavior across surfaces of varying friction, clutter, or compliance. The symmetry of the actuation inherently distributes load, improves fault tolerance, and facilitates rapid self-stabilization under external perturbations or partial actuator failure.  Naturally, this type of design can home distributed perception to enhance object tracking and interaction, especially during large orientation changes. These findings highlight that dynamic symmetry is a principled strategy for coupling morphology, sensing, and control in a single cohesive design space.

Dynamic symmetry also broadens the robot’s functional envelope. Argus’s whole-body actuation supports unconventional behaviors such as rolling, wall climbing, and whole-body loco-manipulation under continuous motions. Its omnidirectional perception allows the robot to track and interact with objects while maintaining high mobility, demonstrating that multifunctionality can emerge directly from a dynamic isotropy-driven design rather than from task-specific appendages or specialized subsystems.

At the same time, our studies reveal meaningful trade-offs. As morphology becomes more symmetric, additional actuators contribute decreasing marginal gains to dynamic isotropy and task performance.  More spatial redundancies with large numbers of legs introduce added mass, mechanical complexity, and computational load, which can reduce energy efficiency and complicate control optimization. These findings point to the potential need for principled co-design of spatial redundancy and dynamic symmetry \cite{dong2023symmetry,ghaffari2022progress}.

Beyond functional advantages, near-extreme dynamic symmetry yields practical engineering benefits. Argus’s modular leg architecture simplifies fabrication, maintenance, replacement, and sim-to-real transfer. Because each foot node can be readily swapped with task-specific attachments, such as high-friction pads for climbing or terrain-adaptive tips for soft substrates, the platform can be customized for new domains without altering its core dynamic symmetry. This modularity positions Argus not only as a scientific demonstrator but as a promising template for domain-specific deployments.

Looking ahead, advances in actuator technology, onboard sensing, and structural materials could expand the operating envelope of dynamic-symmetric robots. Low-gravity environments, in particular, may unlock the full benefits of whole-body actuation and omnidirectional perception for extraterrestrial exploration. Moreover, future Argus variants equipped with multiple degrees of freedom per leg could explore richer loco-manipulation behaviors while preserving the fundamental principles of dynamic symmetry.

More broadly, we envision dynamic symmetry as a guiding principle for a generation of robots that navigate, perceive, and interact with the world with intrinsic robustness and adaptability.  
By demonstrating how extreme dynamic isotropy unifies morphology, perception, and control, Argus provides both a conceptual framework and a
physical platform for rethinking how robots can engage with complex,
unstructured environments.

\section*{Materials and Methods}

\subsection*{Dynamic Symmetry and Dynamic Isotropy}

The isotropy measure is primarily used to describe how directionally uniform a robot's velocity or force transmission capabilities are. It characterizes the manipulability of the end-effector and serves as a key design criterion for robotic manipulators \cite{klein1991spatial,yoshikawa1985manipulability,klein1987dexterity,salisbury1982articulated}. Conventionally, the isotropy is assessed by examining the equality of the singular values of the Jacobian matrix. However, this approach provides only a coarse characterization of the robot’s motion capabilities along orthogonal directions in Cartesian space. To more comprehensively quantify how uniformly the robot can accelerate its center of mass across all spatial directions, we introduced an isotropy measure based on directional acceleration magnitudes.

\mypara{Center of Mass Linear Accelerations Modeling}
Starting from the standard robot dynamics equation,
\begin{equation}
    \boldsymbol{\mathrm{M}}(\boldsymbol{\mathrm{q}})\,\ddot{\boldsymbol{\mathrm{q}}} + \boldsymbol{\mathrm{C}}\bigl(\boldsymbol{\mathrm{q}}, \dot{\boldsymbol{\mathrm{q}}}\bigr)\,\dot{\boldsymbol{\mathrm{q}}} + \boldsymbol{\mathrm{g}}\bigl(\boldsymbol{\mathrm{q}}\bigr) = \boldsymbol{\uptau},
\end{equation}
where \( \boldsymbol{\mathrm{M}}(\boldsymbol{\mathrm{q}}) \in \mathbb{R}^{n \times n} \) is the joint-space inertia matrix, \( \boldsymbol{\mathrm{C}}\bigl(\boldsymbol{\mathrm{q}}, \dot{\boldsymbol{\mathrm{q}}}\bigr) \) represents the Coriolis and centrifugal effects, \( \boldsymbol{\mathrm{g}}\bigl(\boldsymbol{\mathrm{q}}\bigr) \) is the gravity vector, and \( \boldsymbol{\uptau} \in \mathbb{R}^n \) is the joint torque vector.

To focus on the effect of joint configuration on the robot’s acceleration, we analyzed the dynamics under quasi-static or low-velocity conditions (\( \dot{\boldsymbol{\mathrm{q}}} \approx 0 \)) with gravity compensation. Under these assumptions, the velocity-dependent and gravitational terms are negligible, and the dynamics can be approximated by
\begin{equation}
\ddot{\boldsymbol{\mathrm{q}}}
\approx \boldsymbol{\mathrm{M}}^{-1}\bigl(\boldsymbol{\mathrm{q}}\bigr)\,\boldsymbol{\uptau}.
\end{equation}

Let \( J_c(q) \in \mathbb{R}^{3 \times n} \) denote the Jacobian matrix mapping joint velocities to the linear velocity of the center of mass (CoM), that is,
\begin{equation}
\dot{\boldsymbol{\mathrm{x}}}_c
= \boldsymbol{\mathrm{J}}_c\bigl(\boldsymbol{\mathrm{q}}\bigr)\,\dot{\boldsymbol{\mathrm{q}}}.
\end{equation}
Differentiating this relation with respect to time yields the exact expression for the CoM acceleration:
\begin{equation}
\boldsymbol{\mathrm{a}}_c
= \ddot{\boldsymbol{\mathrm{x}}}_c
= \dot{\boldsymbol{\mathrm{J}}}_c\bigl(\boldsymbol{\mathrm{q}}, \dot{\boldsymbol{\mathrm{q}}}\bigr)\,\dot{\boldsymbol{\mathrm{q}}}
+ \boldsymbol{\mathrm{J}}_c\bigl(\boldsymbol{\mathrm{q}}\bigr)\,\ddot{\boldsymbol{\mathrm{q}}},
\end{equation}
where the first term, \( \dot{\boldsymbol{\mathrm{J}}}_c
\bigl(\boldsymbol{\mathrm{q}}, \dot{\boldsymbol{\mathrm{q}}}\bigr)
\,\dot{\boldsymbol{\mathrm{q}}} \), captures velocity-dependent effects, and the second term represents the contribution from joint accelerations.

Under quasi-static or low-velocity motion (\( \dot{\boldsymbol{\mathrm{q}}} \approx \boldsymbol{\mathrm{0}} \)), the first term becomes negligible since it is second-order in velocity. Neglecting these higher-order terms gives the approximation
\begin{equation}
\boldsymbol{\mathrm{a}}_c
\approx \boldsymbol{\mathrm{J}}_c\bigl(\boldsymbol{\mathrm{q}}\bigr)\,\ddot{\boldsymbol{\mathrm{q}}}.
\end{equation}

Substituting the quasi-static joint-space dynamics approximation, we obtain a linear relationship between joint torques and CoM acceleration:
\begin{equation}
\boldsymbol{\mathrm{a}}_c
\approx \boldsymbol{\mathrm{J}}_c\bigl(\boldsymbol{\mathrm{q}}\bigr)\,
\boldsymbol{\mathrm{M}}^{-1}\bigl(\boldsymbol{\mathrm{q}}\bigr)\,\uptau.
\end{equation}

For convenience, we define the configuration-dependent mapping
\begin{equation}
\boldsymbol{\mathrm{A}}\bigl(\boldsymbol{\mathrm{q}}\bigr)
:= \boldsymbol{\mathrm{J}}_c\bigl(\boldsymbol{\mathrm{q}}\bigr)\,
\boldsymbol{\mathrm{M}}^{-1}\bigl(\boldsymbol{\mathrm{q}}\bigr),
\end{equation}
so that the CoM acceleration can be compactly written as
\begin{equation}
\boldsymbol{\mathrm{a}}_c
= \boldsymbol{\mathrm{A}}\bigl(\boldsymbol{\mathrm{q}}\bigr)\,\uptau,
\end{equation}
where \( \boldsymbol{\mathrm{A}}_i\bigl(\boldsymbol{\mathrm{q}}\bigr) \in \mathbb{R}^3 \) denotes the \( i \)-th column of \( \boldsymbol{\mathrm{A}}\bigl(\boldsymbol{\mathrm{q}}\bigr) \), representing the contribution of the \( i \)-th joint torque to the CoM acceleration.

\mypara{Dynamic isotropy}
For each unit direction \( \boldsymbol{\mathrm{u}} \in \mathbb{S}^2 \),the directional sensitivity of actuator \(i\) is
\begin{equation}
    c_i\bigl(\boldsymbol{\mathrm{u}}\bigr)
    = \boldsymbol{\mathrm{u}}^\top
      \boldsymbol{\mathrm{A}}_i\bigl(\boldsymbol{\mathrm{q}}\bigr).
\end{equation}

The maximum achievable CoM acceleration magnitude in direction \(\bm{u}\) is obtained by saturating each actuator at its limit in the direction that maximizes \(\boldsymbol{\mathrm{u}}^\top
\boldsymbol{\mathrm{A}}\bigl(\boldsymbol{\mathrm{q}}\bigr)\,\uptau\):
\begin{equation}
    a_{\max}\bigl(\boldsymbol{\mathrm{u}}\bigr)
    = \max_{\uptau \in \mathcal{T}}
      \boldsymbol{\mathrm{u}}^\top
      \boldsymbol{\mathrm{A}}\bigl(\boldsymbol{\mathrm{q}}\bigr)\,\uptau
    = \sum_{i=1}^{n}
      \big|\,c_i\bigl(\boldsymbol{\mathrm{u}}\bigr)\,\big|
      \,\uptau_i^{\max}.
\end{equation}

We sample a set of directions 
\(\{ \boldsymbol{\mathrm{u}}_k \}_{k=1}^{K}\) 
uniformly distributed on the unit sphere
and compute the directional accelerations 
\(a_{\max}(\boldsymbol{\mathrm{u}}_k)\) of the CoM. 
The dynamic isotropy is then defined as
\begin{equation}
    \upeta\bigl(\boldsymbol{\mathrm{q}}\bigr)
    = \frac{a_{\min}\bigl(\boldsymbol{\mathrm{q}}\bigr)}
           {a_{\max}\bigl(\boldsymbol{\mathrm{q}}\bigr)},
    \qquad
    a_{\min}\bigl(\boldsymbol{\mathrm{q}}\bigr)
    = \min_k a_{\max}\bigl(\boldsymbol{\mathrm{u}}_k\bigr), \quad
    a_{\max}\bigl(\boldsymbol{\mathrm{q}}\bigr)
    = \max_k a_{\max}\bigl(\boldsymbol{\mathrm{u}}_k\bigr),
\end{equation}
where \( \upeta\bigl(\boldsymbol{\mathrm{q}}\bigr) \in [0,1] \).
A perfectly isotropic acceleration capability corresponds to 
\( \upeta\bigl(\boldsymbol{\mathrm{q}}\bigr) = 1 \), 
indicating equal CoM acceleration magnitude in all directions.

\subsection*{Theoretical Analysis of Dynamic Isotropy for Stability, Robustness, and Control Efficiency}

In this section, we show how the dynamic isotropy formulation developed above enables theoretical characterization of stability, robustness, and energy efficiency. All results follow directly from the CoM acceleration mapping
\begin{equation}
    \boldsymbol{\mathrm{a}}_c
    \;=\;
    \boldsymbol{\mathrm{A}}\bigl(\boldsymbol{\mathrm{q}}\bigr)\,\boldsymbol{\uptau},
    \qquad
    \boldsymbol{\mathrm{A}}\bigl(\boldsymbol{\mathrm{q}}\bigr)
    := \boldsymbol{\mathrm{J}}_c\bigl(\boldsymbol{\mathrm{q}}\bigr)\,
       \boldsymbol{\mathrm{M}}^{-1}\bigl(\boldsymbol{\mathrm{q}}\bigr),
\end{equation}
under the quasi-static approximation. Torque limits are modeled as
\begin{equation}
    \uptau_i \in \bigl[-\uptau_i^{\max},\,\uptau_i^{\max}\bigr], 
    \qquad i = 1,\dots,n.
\end{equation}

\mypara{Acceleration reachability and ellipsoidal approximation}
For a configuration \( q \), the set of attainable CoM accelerations is
\begin{equation}
    \mathcal{A}\bigl(\boldsymbol{\mathrm{q}}\bigr)
    := \bigl\{\, \boldsymbol{\mathrm{A}}\bigl(\boldsymbol{\mathrm{q}}\bigr)\,\boldsymbol{\uptau}
    \;\big|\; \boldsymbol{\uptau} \in \boldsymbol{\mathcal{T}} \bigr\},
    \qquad
    \boldsymbol{\mathcal{T}} := \prod_{i=1}^{n}
    \bigl[-\uptau_i^{\max},\,\uptau_i^{\max}\bigr].
\end{equation}
For a direction \( \boldsymbol{\mathrm{u}} \in \mathbb{S}^2 \), the maximum achievable acceleration magnitude is
\begin{equation}
    a_{\max}\bigl(\boldsymbol{\mathrm{u}}\bigr)
    = \max_{\uptau \in \mathcal{T}}
      \boldsymbol{\mathrm{u}}^\top
      \boldsymbol{\mathrm{A}}\bigl(\boldsymbol{\mathrm{q}}\bigr)\,\boldsymbol{\uptau}
    = \sum_{i=1}^{n}
      \bigl|\,
      \boldsymbol{\mathrm{u}}^\top
      \boldsymbol{\mathrm{A}}_i\bigl(\boldsymbol{\mathrm{q}}\bigr)
      \,\bigr|\,
      \uptau_i^{\max}.
\end{equation}
Sampling directions \( \{ \boldsymbol{\mathrm{u}}_k \}_{k=1}^{K} \), we define
\begin{equation}
    a_{\min}\bigl(\boldsymbol{\mathrm{q}}\bigr)
    := \min_k a_{\max}\bigl(\boldsymbol{\mathrm{u}}_k\bigr),
    \qquad
    a_{\max}\bigl(\boldsymbol{\mathrm{q}}\bigr)
    := \max_k a_{\max}\bigl(\boldsymbol{\mathrm{u}}_k\bigr),
\end{equation}
and the dynamic isotropy
\begin{equation}
    \upeta\bigl(\boldsymbol{\mathrm{q}}\bigr)
    := \frac{a_{\min}\bigl(\boldsymbol{\mathrm{q}}\bigr)}
            {a_{\max}\bigl(\boldsymbol{\mathrm{q}}\bigr)}
    \in [0,1].
\end{equation}

To enable theoretical reasoning, we approximate\( \mathcal{A}\bigl(\boldsymbol{\mathrm{q}}\bigr) \) 
by its minimum-volume ellipsoid (Löwner--John ellipsoid) with shape matrix 
\( \boldsymbol{\mathrm{Q}}\bigl(\boldsymbol{\mathrm{q}}\bigr) \succ 0 \). 
Let \( \lambda_1 \ge \lambda_2 \ge \lambda_3 > 0 \) be the eigenvalues of 
\( \boldsymbol{\mathrm{Q}}\bigl(\boldsymbol{\mathrm{q}}\bigr) \). Then
\begin{equation}
    a_{\max}\bigl(\boldsymbol{\mathrm{q}}\bigr)
    \approx \sqrt{\lambda_1},
    \qquad
    a_{\min}\bigl(\boldsymbol{\mathrm{q}}\bigr)
    \approx \sqrt{\lambda_3},
\end{equation}
and dynamic isotropy is inversely related to the condition number:
\begin{equation}
    \upeta\bigl(\boldsymbol{\mathrm{q}}\bigr)
    \approx \frac{\sqrt{\lambda_3}}{\sqrt{\lambda_1}}
    = \frac{1}{\sqrt{\kappa\bigl(\boldsymbol{\mathrm{Q}}\bigl(\boldsymbol{\mathrm{q}}\bigr)\bigr)}},
    \qquad
    \kappa\bigl(\boldsymbol{\mathrm{Q}}\bigl(\boldsymbol{\mathrm{q}}\bigr)\bigr)
    := \frac{\lambda_1}{\lambda_3}.
\end{equation}
Thus, high dynamic isotropy corresponds to a well-conditioned acceleration-map ellipsoid.

\mypara{Stability margins}
To analyze closed-loop stability, we consider the set of all feasible CoM accelerations at configuration $\boldsymbol{\mathrm{q}}$. Under actuator limits, this set can be approximated by the minimum-volume ellipsoid
\begin{equation}
\boldsymbol{\mathrm{E}}\bigl(\boldsymbol{\mathrm{q}}\bigr)
=
\left\{\,
\boldsymbol{\mathrm{a}} \in \mathbb{R}^3
\;\middle|\;
\boldsymbol{\mathrm{a}}^\top
\boldsymbol{\mathrm{Q}}^{-1}\bigl(\boldsymbol{\mathrm{q}}\bigr)\,
\boldsymbol{\mathrm{a}}
\le 1
\right\},
\qquad
\boldsymbol{\mathrm{Q}}\bigl(\boldsymbol{\mathrm{q}}\bigr) \succ 0,
\label{eq:ellipsoid-def}
\end{equation}
whose shape matrix $\boldsymbol{\mathrm{Q}}\bigl(\boldsymbol{\mathrm{q}}\bigr)$ captures the distribution of attainable accelerations in all directions.

Suppose a controller requires a corrective CoM acceleration $\boldsymbol{\mathrm{a}}_{\mathrm{req}} $ to reject a disturbance or stabilize the system. For the robot to remain stable, this commanded acceleration must lie within the feasible acceleration set $\boldsymbol{\mathrm{E}}\bigl(\boldsymbol{\mathrm{q}}\bigr)$. 
Thus, feasibility requires
\begin{equation}
\boldsymbol{\mathrm{a}}_{\mathrm{req}} 
\in 
\boldsymbol{\mathrm{E}}\bigl(\boldsymbol{\mathrm{q}}\bigr).
\end{equation}

Substituting Eq.~\eqref{eq:ellipsoid-def} gives the explicit feasibility condition
\begin{equation}
\boldsymbol{\mathrm{a}}_{\mathrm{req}}^\top
\boldsymbol{\mathrm{Q}}^{-1}\bigl(\boldsymbol{\mathrm{q}}\bigr)\,
\boldsymbol{\mathrm{a}}_{\mathrm{req}}
\;\le\; 1.
\label{eq:stability-feasibility}
\end{equation}
Geometrically, $\boldsymbol{\mathrm{Q}}^{-1}\bigl(\boldsymbol{\mathrm{q}}\bigr)$ defines the ``cost'' of producing acceleration in each direction: accelerations aligned with weak axes of the ellipsoid (small eigenvalues of $\boldsymbol{\mathrm{Q}}$) incur larger quadratic cost.

To quantify how close the system is to violating feasibility, we define the stability margin as
\begin{equation}
\mathrm{margin}\bigl(\boldsymbol{\mathrm{q}}\bigr)
= 1 -
\boldsymbol{\mathrm{a}}_{\mathrm{req}}^\top
\boldsymbol{\mathrm{Q}}^{-1}\bigl(\boldsymbol{\mathrm{q}}\bigr)\,
\boldsymbol{\mathrm{a}}_{\mathrm{req}},
\label{eq:stability-margin}
\end{equation}
which equals $1$ when no corrective acceleration is required and decreases to $0$ when the commanded acceleration reaches the boundary of the feasible ellipsoid.
A larger margin indicates greater tolerance to disturbances and improved closed-loop stability. To ensure feasibility, the required corrective acceleration must lie inside the ellipsoidal feasible set \( \boldsymbol{\mathrm{E}}\bigl(\boldsymbol{\mathrm{q}}\bigr) \), 
which implies 
\( \|\boldsymbol{\mathrm{a}}_{\mathrm{req}}\| \le \sqrt{\lambda_3} \) 
along the weakest principal axis. 
The absolute disturbance rejection capability, therefore, scales with \( \sqrt{\lambda_3} \), 
whereas dynamic isotropy \( \upeta \approx \sqrt{\lambda_3/\lambda_1} \)  controls how uniformly this capability is distributed across directions. Robots with higher dynamic isotropy avoid having disproportionately weak axes, reducing directional vulnerabilities and enabling orientation-invariant stabilization. Even when $\lambda_1$ varies across morphologies, higher $\upeta$ ensures that $\lambda_3$ does not collapse relative to $\lambda_1$, preserving larger stability margins under arbitrary disturbance directions.

\mypara{Disturbance rejection and robustness}
External forces \( \boldsymbol{\mathrm{F}}_{\mathrm{ext}} \) induce a required CoM acceleration 
\( \boldsymbol{\mathrm{a}}_{\mathrm{req}} 
= \boldsymbol{\mathrm{M}}^{-1}\bigl(\boldsymbol{\mathrm{q}}\bigr)\,\boldsymbol{\mathrm{F}}_{\mathrm{ext}} \). 
Feasibility under worst-case alignment occurs when the induced acceleration points along the weakest principal axis of the feasible ellipsoid 
\( \boldsymbol{\mathrm{E}}\bigl(\boldsymbol{\mathrm{q}}\bigr) \), meaning
\begin{equation}
    \|\boldsymbol{\mathrm{F}}_{\mathrm{ext}}\|_{\max}
    \;\propto\;
    \sqrt{\lambda_{3}}.
\end{equation}
Since dynamic isotropy satisfies
\( \upeta\bigl(\boldsymbol{\mathrm{q}}\bigr)=\sqrt{\lambda_3/\lambda_1} \), 
we could also express this as
\begin{equation}
    \|\boldsymbol{\mathrm{F}}_{\mathrm{ext}}\|_{\max}
    \;\propto\;
    \upeta\bigl(\boldsymbol{\mathrm{q}}\bigr)\,\sqrt{\lambda_1},
\end{equation}
highlighting that $\lambda_3$ governs the absolute disturbance rejection capability whereas $\upeta(\boldsymbol{\mathrm{q}})$ quantifies its directional uniformity. Higher dynamic isotropy prevents the emergence of dynamically weak directions, ensuring that the minimum achievable acceleration $\sqrt{\lambda_3}$ does not collapse relative to the maximum axis $\sqrt{\lambda_1}$. As a result, robots with higher dynamic isotropy maintain more uniform resilience to terrain irregularities, collisions, uneven contacts, or actuator outages, consistent with our empirical findings in cluttered environments and friction-varying surfaces.

\mypara{Control efficiency}
Producing a desired CoM acceleration 
\( \boldsymbol{\mathrm{a}}_{\mathrm{des}} \) requires joint torques \( \uptau \) satisfying 
\( \boldsymbol{\mathrm{a}}_{\mathrm{des}} = \boldsymbol{\mathrm{A}}\bigl(\boldsymbol{\mathrm{q}}\bigr)\,\uptau \). 
The minimum-energy torque command solves the quadratic program
\begin{equation}
    \uptau^\star
    = \arg\min_{\uptau}\;\|\uptau\|_2^2
    \quad\text{s.t.}\quad
    \boldsymbol{\mathrm{A}}\bigl(\boldsymbol{\mathrm{q}}\bigr)\,\uptau
    = \boldsymbol{\mathrm{a}}_{\mathrm{des}}.
\end{equation}
The optimal solution is given by the Moore--Penrose pseudoinverse:
\begin{equation}
    \uptau^\star
    = \boldsymbol{\mathrm{A}}^{\dagger}\bigl(\boldsymbol{\mathrm{q}}\bigr)\,
      \boldsymbol{\mathrm{a}}_{\mathrm{des}},
    \qquad
    \|\uptau^\star\|_2^2
    =
    \boldsymbol{\mathrm{a}}_{\mathrm{des}}^\top
    \left(
    \boldsymbol{\mathrm{A}}\bigl(\boldsymbol{\mathrm{q}}\bigr)\,
    \boldsymbol{\mathrm{A}}^\top\bigl(\boldsymbol{\mathrm{q}}\bigr)
    \right)^{-1}
    \boldsymbol{\mathrm{a}}_{\mathrm{des}}.
\end{equation}

To connect this energy requirement to the feasible acceleration ellipsoid, similar as above, we approximated $\boldsymbol{\mathrm{A}}\bigl(\boldsymbol{\mathrm{q}}\bigr)$ by its minimum-volume enclosing ellipsoid with shape matrix $\boldsymbol{\mathrm{Q}}\bigl(\boldsymbol{\mathrm{q}}\bigr)\succ 0$. Since \( \boldsymbol{\mathrm{A}}\bigl(\boldsymbol{\mathrm{q}}\bigr)
   \boldsymbol{\mathrm{A}}^\top\bigl(\boldsymbol{\mathrm{q}}\bigr)
   \approx \boldsymbol{\mathrm{Q}}\bigl(\boldsymbol{\mathrm{q}}\bigr) \), the minimum required energy becomes
\begin{equation}
    \|\uptau^\star\|_2^2
    \;\approx\;
    \boldsymbol{\mathrm{a}}_{\mathrm{des}}^\top
    \boldsymbol{\mathrm{Q}}^{-1}\bigl(\boldsymbol{\mathrm{q}}\bigr)\,
    \boldsymbol{\mathrm{a}}_{\mathrm{des}}.
\end{equation}
Thus, control energy is proportional to the ellipsoidal norm of the desired acceleration. If $\boldsymbol{\mathrm{a}}_{\mathrm{des}}$ is decomposed along the principal axes of $\boldsymbol{\mathrm{Q}}\bigl(\boldsymbol{\mathrm{q}}\bigr)$, we obtained
\begin{equation}
    \|\uptau^\star\|_2^2
    \approx
    \frac{\alpha_1^2}{\lambda_1}
    + \frac{\alpha_2^2}{\lambda_2}
    + \frac{\alpha_3^2}{\lambda_3},
\end{equation}
where $\alpha_i$ is the projection of $a_{\mathrm{des}}$ onto eigenvector $i$. Control becomes highly imbalanced when $\lambda_3$ is much smaller than $\lambda_1$, because accelerations aligned with the weakest axis require disproportionately larger torques.

Dynamic isotropy quantifies this imbalance through $\upeta\bigl(\boldsymbol{\mathrm{q}}\bigr) \approx \sqrt{\lambda_3/\lambda_1}$. A higher isotropy value reduces the ratio $\lambda_1/\lambda_3$, ensuring that no direction demands excessive joint effort relative to others. Thus, high dynamic isotropy directly improves control efficiency by reducing the worst–case and directional variability of the actuation effort needed to realize CoM accelerations of a given magnitude. This analytic prediction aligns with our empirical findings: robots with higher dynamic isotropy consistently required less total actuation effort to achieve the same locomotion velocities and maneuvering agility, especially when subjected to orientation changes, uneven surfaces, or partial actuator load.

\mypara{Summary}
Dynamic isotropy provides a quantitative and mechanistic link between dynamic symmetry and whole-body capability. High dynamic isotropy ensures that the robot’s acceleration map is well conditioned with orientation-invariant stability margins, uniform robustness to external disturbances, and balanced control efforts. These analytical predictions can help understand the empirical advantages observed in dynamically symmetric Argus variants and the trend toward the theoretical extreme of dynamic isotropy, including orientation-invariant locomotion, agile recovery behavior, energy efficiency, and multifunctional behaviors across a wide range of environments.

\mypara{Dynamic symmetry vs redundancy}
Here we showed that redundancy primarily rescales the feasible-acceleration ellipsoid, whereas dynamic symmetry determines its shape, and therefore governs the stability, robustness, and efficiency results derived above in this work. Recall that the acceleration ellipsoid is characterized by
\begin{equation}
\boldsymbol{\mathrm{Q}}\bigl(\boldsymbol{\mathrm{q}}\bigr)
=
\sum_{i=1}^n 
\uptau_i^{\max\,2}\,
\boldsymbol{\mathrm{A}}_i\bigl(\boldsymbol{\mathrm{q}}\bigr)
\boldsymbol{\mathrm{A}}_i\bigl(\boldsymbol{\mathrm{q}}\bigr)^\top,
\qquad 
\boldsymbol{\mathrm{Q}}\bigl(\boldsymbol{\mathrm{q}}\bigr)\succ 0,
 \end{equation}
whose eigenvalues satisfy 
\(
\lambda_1\ge \lambda_2\ge \lambda_3>0.
\)
The dynamic isotropy is
\begin{equation}
\upeta\bigl(\boldsymbol{\mathrm{q}}\bigr) 
\approx 
\sqrt{\frac{\lambda_3}{\lambda_1}},
\end{equation}
which depends only on the relative spread of the eigenvalues. Multiplying $ \boldsymbol{\mathrm{Q}}\bigl(\boldsymbol{\mathrm{q}}\bigr)$ by any positive scalar $c$ scales all eigenvalues equally, leaving $\upeta\bigl(\boldsymbol{\mathrm{q}}\bigr)$ invariant. Thus, isotropy is insensitive to uniform expansion of the acceleration set.

Suppose $k$ additional actuators are added whose directions repeat an existing direction $A_j(q)$.  The new matrix is
\begin{equation}
\boldsymbol{\mathrm{Q}}'\bigl(\boldsymbol{\mathrm{q}}\bigr)
=
\boldsymbol{\mathrm{Q}}\bigl(\boldsymbol{\mathrm{q}}\bigr)
+
\sum_{\ell=1}^{k}
\uptau_{\ell}^{\max\,2}\,
\boldsymbol{\mathrm{A}}_j\bigl(\boldsymbol{\mathrm{q}}\bigr)
\boldsymbol{\mathrm{A}}_j\bigl(\boldsymbol{\mathrm{q}}\bigr)^\top
=
\boldsymbol{\mathrm{Q}}\bigl(\boldsymbol{\mathrm{q}}\bigr)
+
c\,
\boldsymbol{\mathrm{A}}_j\bigl(\boldsymbol{\mathrm{q}}\bigr)
\boldsymbol{\mathrm{A}}_j\bigl(\boldsymbol{\mathrm{q}}\bigr)^\top.
\end{equation}
This update affects only the eigenvalue associated with $A_j(q)$, increasing $\lambda_1$ while leaving $\lambda_2$ and $\lambda_3$ almost unchanged:
\begin{equation}
\lambda_1'=\lambda_1+\Delta,\qquad 
\lambda_2'\approx\lambda_2,\quad 
\lambda_3'\approx\lambda_3.
\end{equation}
Hence
\begin{equation}
\upeta'\bigl(\boldsymbol{\mathrm{q}}\bigr) 
\approx 
\sqrt{\frac{\lambda_3'}{\lambda_1'}} 
\approx 
\sqrt{\frac{\lambda_3}{\lambda_1 + \Delta}}
\le 
\upeta\bigl(\boldsymbol{\mathrm{q}}\bigr).
\end{equation}
Redundancy alone therefore could reduce isotropy in this extreme case. It cannot explain the observed improvement in worst-case acceleration, stability margin, or robustness. Similarly, increases in actuator count that do not diversify the actuation directions expand the acceleration ellipsoid anisotropically, improving best-case acceleration but leaving the weakest direction unchanged. Even when such additions are distributed evenly for the actuation directions, such redundancy only changes the scale of the ellipsoid, not its overall shape such as uniformity. In contrast, as we had shown above, more dynamic symmetric designs can reshape the full matrix $ \boldsymbol{\mathrm{Q}}\bigl(\boldsymbol{\mathrm{q}}\bigr)$ to produce a more spherical ellipsoid and increase the dynamic isotropy score.

\subsection*{Mechanical Design}
The 20-legs version weighs \SI{23.4}{\kilogram} and the overall span of the structure ranges from \SI{0.95}{\meter} in its fully retracted state to \SI{1.37}{\meter} in its fully extended state(Fig. S\ref{fig:extended_argus_sizing}), corresponding to a leg acutation range of $0.21\si{\metre}$. The home position is defined at the midpoint of the leg's range, allowing for \SI{0.105}{\meter} of actuation in both the positive and negative directions. Each custom actuator delivers up to 375N peak thrust at 1m/s while weighing only 0.62 kg, which produces 2 to 13 times maximum thrust per actuator mass than all previous solutions (Table S\ref{table:robot_actuators}) where linear actuators have been used in prior tensegrity or spherical robots. A
time-of-flight (ToF) camera is equipped at the tip of each leg, providing omnidirectional range sensing(Fig.~\ref{figure_1}B) that remains unobstructed by leg motion.
\mypara{Modular leg} To achieve the high-speed and high-force linear actuation required for dynamic and responsive locomotion, we employ a cable-driven mechanism that converts rotational motion into linear actuation (Fig.~\ref{figure_1}A). Each leg costs approximately \$300 and exceeds the off-the-shelf linear actuators in both speed and thrust force. The module consists of a quasi-direct-drive motor (RoboStride 02) housed in a custom motor mount. A rotary drum disk directly coupled to the motor guides a single rope (9KM DWLIFE, 2.3mm braided Kevlar) routed in both directions and anchored at the two ends of the moving assembly. The moving assembly comprises two triangular end plates connected by three carbon fiber rods ($\varnothing6\,\si{mm} \times 300\,\si{mm}$). Each rod was held by two LM6UU linear bearings ($6/12/19\,\si{mm}$ bore/outer diameter/length) embedded in the motor mount. These bearings constrained the motion of the carbon fiber rods to a single axis, allowing the moving assembly to slide linearly along the actuator axis as the cable on either end is tensioned or released. The outer end-plate was connected to a ball-shaped foot node with a radius of $0.06\,\si{\meter}$, secured in place using screws. The leg modules were interconnected using short carbon fiber rods ($\varnothing6\,\si{mm} \times 150\,\si{mm}$), forming the edges of the dodecahedron. 
Additional carbon fiber rods ($\varnothing6\,\si{mm} \times 300\,\si{mm}$) connected each leg module to the central support structure, passing through the center of the inner end plate and secured with a bushing ($6/8/10\,\si{mm}$ bore/outer diameter/length). These rods provide additional support for each foot, improving overall structural rigidity.

\mypara{Sensors} A time-of-flight (ToF) camera (MaixSense-A010) was attached to the tip of the foot node such that the sensor can capture the view in the direction of actuation without occlusion. The ToF camera has a maximum resolution of 100\,$\times$\,100 pixels, a field of view \SI{70}{\degree}$\times$\SI{60}{\degree}, a maximum range of 2.5 \SI{m}, and a maximum frame rate of 20Hz. We designed the concave shape on the tip of the foot node to ensure enough opening for the field of view of the ToF sensor. The twenty ToF cameras were evenly distributed at the end position of each foot node such that the robot had full coverage of the environment around its body.
In simulation, the ToF camera was modeled via ray casting over a 3D mesh representation of the scene. Rays were emitted from the origin of the foot node and intersected with scene geometry, including a planar mesh representing the ground and cuboid meshes representing objects. The resulting intersection points were computed to emulate the sensor's depth measurements. Each perception unit simulated a fixed spatial resolution of 5×5 rays evenly distributed based on the field of view.  Only depth measurements within a maximum range of \SI{1.5}{\meter}, measured from the tip of the foot node, were retained and converted into point cloud data. Rays returning distances beyond this threshold were masked with zeros. In physical experiments, we downsampled the depth image and zero out the distance that exceeds the \SI{1.5}{\meter} maximum range. The distance threshold was chosen based on the acceptable noise level with respect to the perception distance. The resolution was chosen by balancing the perceived information, the communication bandwidth, and the delays of the signal synchronizations based on our onboard power and computing unit.

\mypara{Electronics} To preserve mass symmetry, the electronics and battery were mounted on opposite faces of the dodecahedral structure. The controller stack was placed on one face of Argus, whereas two 6-cell, \SI{5200}{\milli\ampere\hour} lithium-polymer batteries are positioned on the other side of Argus. The two batteries were connected in series to power all twenty actuators. The robot was controlled using an onboard computer (NVIDIA, Jetson Orin Nano) powered by a USB power bank. The inertial measurement unit (IMU) (SYD Dynamics, TM171) was mounted below the Jetson computer, providing orientation and angular velocity estimates. To enable synchronized streaming of all 20 Tof camera signals, we included an additional mini PC (GMKtec NucBox G2 PLUS) to stream 14 sensors' readings and integrate with the remaining six sensors accessed from Jetson through Ethernet. All the electric cables were tightened on the frames of the base body.

\mypara{Fabrication} Most structural elements were 3D-printed using polylactic acid (PLA), including the center support, motor connectors, carbon fiber rod mounts, actuator end plates, and the electronics and battery holder. Carbon fiber rods, linear bearings, and other electronic parts were sourced off the shelf. Each actuator was interconnected with its three neighboring actuators using carbon fiber rods, clamped together with screw connections to form the outer edges of the structure. To further improve rigidity, the midpoints of the edges on each pentagonal face were connected by tensioned ropes arranged in a pentagram pattern.

\subsection*{Policy Training, Tasks and Evaluations}

\mypara{Policy training} We modeled each control task as a Partially Observable Markov Decision Process (POMDP)~\cite{kaelbling1998planning}, where the control policy receives partial and noisy observations from onboard sensors and must infer its underlying state in order to make decisions. Let $\boldsymbol{\mathrm{\pi}}_{\uptheta}
\bigl(\boldsymbol{\mathrm{a}}_{\mathrm{t}} \mid \boldsymbol{\mathrm{o}}_{\mathrm{t}}\bigr)$ denote a policy parameterized by $\uptheta$, mapping observations $\boldsymbol{\mathrm{o}}_{\mathrm{t}}$ to actions $\boldsymbol{\mathrm{a}}_{\mathrm{t}}$. The objective is to maximize the expected cumulative reward:

\begin{equation}
\max_{\uptheta} \; 
\mathbb{E}_{\boldsymbol{\mathrm{\pi}}_{\uptheta}}
\left[
\sum_{\mathrm{t}=0}^{T}
\upgamma^{\mathrm{t}} \,
\mathrm{r}\bigl(\boldsymbol{\mathrm{s}}_{\mathrm{t}}, \boldsymbol{\mathrm{a}}_{\mathrm{t}}\bigr)
\right]
\label{eq:rl_objective}
\end{equation}

where $\mathrm{r}\bigl(\boldsymbol{\mathrm{s}}_{\mathrm{t}}, \boldsymbol{\mathrm{a}}_{\mathrm{t}}\bigr)$ and $\boldsymbol{\mathrm{s}}_{\mathrm{t}}$ is the reward and true state at time $t$, and $\upgamma \in (0,1]$ is the discount factor. We trained the control policy in Isaac Gym~\cite{makoviychuk2021isaac} using the Proximal Policy Optimization (PPO) algorithm~\cite{schulman2017proximal}, which improved policy performance through stable updates by optimizing the clipped surrogate objective:

\begin{equation}
\boldsymbol{\mathrm{L}}^{\mathrm{CLIP}}(\uptheta)
=
\hat{\mathbb{E}}_{\mathrm{t}}
\left[
\min \left(
\mathrm{r}_{\mathrm{t}}(\uptheta)\,\hat{\boldsymbol{\mathrm{A}}}_{\mathrm{t}},\;
\mathrm{clip}\bigl(
\mathrm{r}_{\mathrm{t}}(\uptheta),\,
1 - \epsilon,\,
1 + \epsilon
\bigr)\,
\hat{\boldsymbol{\mathrm{A}}}_{\mathrm{t}}
\right)
\right],
\label{eq:ppo_loss}
\end{equation}

where $\hat{\mathbb{E}}_{\mathrm{t}}$ denotes the empirical average over a finite batch of collected timesteps, 
$ \mathrm{r}_{\mathrm{t}}(\uptheta) 
= \frac{
\boldsymbol{\mathrm{\pi}}_{\uptheta}
\bigl(\boldsymbol{\mathrm{a}}_{\mathrm{t}} \mid \boldsymbol{\mathrm{o}}_{\mathrm{t}}\bigr)
}{
\boldsymbol{\mathrm{\pi}}_{\uptheta_{\mathrm{old}}}
\bigl(\boldsymbol{\mathrm{a}}_{\mathrm{t}} \mid \boldsymbol{\mathrm{o}}_{\mathrm{t}}\bigr)
} $
is the probability ratio between the new and old policies, 
$\hat{\boldsymbol{\mathrm{A}}}_{\mathrm{t}}$ is the estimated advantage function, 
and $\epsilon$ is a clipping hyperparameter. Task configurations and evaluation metrics are detailed below.

\mypara{Baseline locomotion on flat terrain} During training and evaluation, 8192 Argus robots were initialized on flat ground with random orientations and were commanded to follow a target velocity uniformly sampled from $-0.8$ to $0.8$ \SI{}{\meter/\second} in the horizontal plane. A total of 8192 evaluations were conducted, and linear velocity error was measured at \SI{3}{\second} after the start of each trial.

\mypara{Self-stabilization} During training, 8192 Argus robots were initialized \SI{0.5}{\meter} above the ground with randomized orientations and zero commanded velocity. At initialization, each robot received a perturbation: a randomly sampled linear velocity in the range of $-2$ to $2$ \SI{}{\meter/\second} and angular velocity in the range of $-2$ to $2$ \SI{}{\radian/\second}, applied across all three spatial axes. A total of 8192 evaluations were conducted. A. stabilization distance was measured as the horizontal distance from the point of first ground contact until the robot comes to rest or until the end of the episode.

\mypara{Carry object} Argus was trained under the same conditions as the baseline locomotion task, with additional randomizations of base mass, inertia, and center of mass position. A total of 8192 evaluations were conducted, with a random mass between 0 and \si{40}{kg} added to the baselink. During evaluation, Argus was commanded to move at a constant velocity of \SI{0.8}{\meter/\second} for \SI{10}{\second}. A success rate was computed as the fraction of trials in which the robot traveled at least 50\% of the commanded distance. For real-world testing, a weighted pouch was attached to the surface of the robot, and the robot was commanded to move at \SI{0.6}{\meter/\second} in the forward (+x) direction.

\mypara{Locomotion with leg failure} In addition to the baseline locomotion task, the legs of Argus were randomly disabled; each leg had a $10\%$ chance to be disabled. Argus was evaluated under a constant velocity command of \SI{0.8}{\meter/\second} for a duration of \SI{10}{\second}. A success rate was computed as the fraction of trials in which the robot travels at least 50\% of the commanded distance. A total of 8192 evaluations were conducted.  For physical experiments, we randomized the number of legs disabled and the orientation of the robot, and commanded the robot to move at \SI{0.6}{\meter/\second} in the forward (+x) direction.

\mypara{Discrete terrain traversal} In addition to flat-ground locomotion, Argus was trained on discrete obstacle terrain, where rectangular block obstacles up to \SI{0.1}{\meter} in height were placed randomly. During training, the obstacle height was gradually increased based on the robot's tracking performance. For evaluation, a total of 8192 evaluations were sampled, the robot moved at a commanded velocity of \SI{0.8}{\meter/\second} for \SI{10}{\second}. It was considered successful if it traversed half of the commanded distance (\SI{4}{\meter}), and the success rate was the fraction of trials that met this criterion. In physical experiments, the trial was considered successful if the robot traveled through all the discrete blocks.

\mypara{Climb up wall at low gravity} Argus was initialized between two parallel rough walls spaced \SI{1}{\meter} apart. It was tasked with climbing the walls as quickly as possible, with part of its weight supported upward to simulate low gravity. During training, 4096 robots were initialized at random heights, positions, and orientations between the walls. The base support force was gradually reduced based on the average velocity of the population, simulating from $8.2\%$ Earth's gravity to the Moon's gravity ($16.6\%$ Earth's gravity). Linear velocity along the vertical direction was measured for this task. For evaluation, 4096 robots were initialized on the ground between the same \SI{1}{\meter} parallel walls. A trial is considered successful if the robot achieves a vertical velocity of at least \SI{0.08}{\meter\per\second} by the end of the \SI{5}{\second} trial. For the physical experiments, we attached a \SI{19.5}{\kilogram} counterweight to the robot using a pulley system with two rolling wheels to simulate lunar gravity. In each trial, the robot was initialized in the same starting pose.

\mypara{Object tracking} The object was randomly initialized around Argus with an offset of \SI{1.5}{\meter}. The object y-axis was parallel with the direction of commanded linear velocity. The linear velocity command amplitude remained the same across the episode of the environment and was randomly sampled within the range of $0.5$ to $0.8$ \si{\meter/\second}. The reward encouraged the robot to match the object's velocity. The performance was evaluated over 7168 environments for ten seconds of tracking with random initial robot orientation. The object's velocity was randomly sampled, consistent with the training setup. The success rate was measured based on the distance of the robot and object within the perceptual range of the robot. We used a maximum distance of \SI{2.5}{\meter} in the first five seconds of tracking as the criterion in the success rate evaluation. In physical experiments, the trial was considered successful if the robot continuously tracked the object to the end of the object’s traveled distance.

\mypara{Object pushing} The object was initialized in front of Argus with \SI{0.8}{\meter} offset away from the robot's position. The commanded velocity was randomly sampled within the range of $-40\degree$ to $40\degree$ with the same normal amplitude of \SI{0.6}{\meter/\second}. The performance was evaluated over 7168 environments for ten seconds of pushing with random initial orientation. A trial was considered successful if the robot-object distance remained less than \SI{1.6}{\meter} throughout the \SI{10}{\second} of the pushing, and the robot did not surpass the object with a minimal robot-object distance of \SI{0.5}{\meter} in the commanded pushing direction. In the real-world experiment, a trial was considered successful if the robot continuously pushed the object until it reached the end of the experimental space.

\mypara{Two-Stage learning from ToF observations} For the perception-related task, computing the large amount of distance information to simulate 20 ToF cameras is slow with large-scale parallel training in the Isaac-gym simulator. To facilitate training, we adopted a two-stage training pipeline. In the first stage, we trained the policy given privileged information about the object states using reinforcement learning, optimizing for the expected cumulative reward of the task. Such information is not available during real-world deployment. In the second stage, we rolled out the locomotion policy and collected the point cloud and object state pairs. We supervised the training of a point cloud encoder to predict the object states, including the object x and y direction velocity in the object-tracking task and the object orientation in the object-pushing task. The point cloud encoder consisted of two PointNet \cite{qi2017pointnet} layers to encode the input contact points to a global feature vector with the dimension of $1 \times 1024$. This feature vector was then passed through three fully connected layers to predict the object states. To deploy the policy using point cloud observations directly from the ToF cameras, the encoder predicted the object states, which were then concatenated with other proprioceptive observations and fed into the locomotion policy. Given a point cloud observation $\boldsymbol{\mathrm{P}}_i$, we trained the point cloud encoder $\boldsymbol{\mathrm{f}}_{\mathrm{p}} $, parameterized by $\uptheta_{\mathrm{p}}$, to output the corresponding object states $\hat{\boldsymbol{\mathrm{O}}}_i$. We optimized the following loss function $\mathcal{L}_{\mathrm{MSE}}$ based on the mean squared error (MSE) loss:
\begin{equation}
    \min_{\uptheta_{\mathrm{p}}}
    \sum_{i}
    \mathcal{L}_{\mathrm{MSE}}
    \bigl(
    \hat{\boldsymbol{\mathrm{O}}}_i,\,
    \boldsymbol{\mathrm{O}}_i
    \bigr),
    \qquad
    \hat{\boldsymbol{\mathrm{O}}}_i
    =
    \boldsymbol{\mathrm{f}}_{\mathrm{p}}
    \bigl(
    \boldsymbol{\mathrm{P}}_i
    \bigr)
\end{equation}

\mypara{Observations} The proprioceptive observation included angular velocity $\mathbf{v}_b$, joint position $\mathbf{q}$, joint velocity $\dot{\mathbf{q}}$, previous action $\mathbf{a}_{t-1}$, and base rotation matrix $\mathbf{R}$. Up to $10\%$ of noise was applied to each observation. The detailed list of observations and states for each task was listed in the Table S\ref{table:obs_state}. In physical experiments, the angular velocity and base rotation were estimated from the IMU. The linear joint position and velocity were derived from motor encoder readings, scaled by the drum radius. We stacked three frames of the observation for the locomotion policy. We used a single frame observation in the object tracking, object pushing, and wall climbing tasks.

\mypara{Actions} The policy's actions $\mathbf{a_t}$ were smoothed using a first-order exponential low-pass filter ($\mathbf{a}^*_{t}={\alpha}\mathbf{a}_t+(1-\alpha)\mathbf{a}_{t-1}^*$, where $\alpha$ is randomized each step within the range of $0.5$ to $0.9$). The smoothed action was then converted to target joint positions ($\mathbf{q}^*=\mathbf{q}_0+k_q\mathbf{a}^*_{t}$) at \SI{25}{\hertz}. Here, $\mathbf{q}_0$ is the default joint positions, and $k_q$ is a scaling factor for the action. For the wall-climbing task, action smoothing was disabled to allow rapid action changes required for fast and responsive behavior. The target joint positions were transformed into torques via a position PD controller, operating at \SI{200}{\hertz} in simulation. For physical experiments, the position PD controller operates asynchronously at \SI{200}{\hertz}. The empirically determined PD gains were $\SI[inter-unit-product=\cdot\mkern-1mu]{10}{\newton\meter/\radian}$ and $\SI[inter-unit-product=\cdot\mkern-1mu]{0.6}{\newton\meter\second/\radian}$.

\mypara{Reward} The reward functions for the baseline velocity tracking task (Table S\ref{table:baseline_reward}) include tracking and regularization terms. 
\(
\boldsymbol{\upphi}\bigl(\boldsymbol{\uppsi}, \boldsymbol{\mathrm{w}}\bigr)
:= \exp\!\left(\sum_{i=1}^{n} \boldsymbol{\mathrm{w}}_{i}\boldsymbol{\uppsi}_{i}^{2}\right)
\)
is the exponential of the weighted sum of squared quantities $\uppsi$, where $\mathrm{w}$ is a corresponding scaling factor. $\mathbf{v}_b$ and $\mathbf{g}_b$ represent the base linear velocity and the projected gravity vector, respectively. $\mathbf{q}_{\min}$ and $\mathbf{q}_{\max}$ define the joint position limits. $\mathbf{a}$ and $\dot{\mathbf{a}}$ refer to the action and action rate, respectively. $\mathbf{\uptau}$ denotes the joint actuation force, $\dot{\mathbf{q}}$ the joint velocity, $\ddot{\mathbf{q}}$ the joint acceleration, and $\mathbf{F}_{cz}$ the contact force in the vertical direction. $\mathbf{I}$ is a boolean indicator of foot contact, and $G$ is the robot's total gravitational force.
For the object tracking task, the object’s velocity is defined the same as $\mathbf{v}_b^*$ and used in the linear velocity reward. In the object pushing task, two additional rewards are included. One encourages the alignment of the object’s y-axis direction with the commanded velocity, and the other encourages the robot to move closer to the object. The term $\boldsymbol{\mathrm{y}}_{\mathrm{obj}}$ denotes the normalized y-axis vector of the object. The commanded pushing velocity is represented by $\boldsymbol{\mathrm{v}}_{\mathrm{command}}$,  $\boldsymbol{\mathrm{p}}_{\mathrm{robot}}$ and $\boldsymbol{\mathrm{p}}_{\mathrm{object}}$ refer to the positions of the robot and the object, respectively.

\mypara{Sim-to-real transfer} During the locomotion policy training, domain randomization (Table S\ref{table:randomization}) was applied to cover possible variations of real-world dynamics, including loose cable or uneven cable tension, leg deformation, and friction in foot nodes and linear bearings. To transfer the trained policy to the real world, we commanded the five-foot node to track a sinusoidal trajectory for both the sim and real robot. We tuned motor parameters ($\mathrm{k}_\mathrm{p}$, $\mathrm{k}_\mathrm{d}$) and the torque ratio to match the motor position and velocity responses between the sim and the real. In the simulation, a \SI{20}{\milli\second} action delay was added to match the hardware latency. We also found that battery level could affect motor torque and influence the overall sim-to-real transfer. In our physical experiments, we found no major influence of policy transfer under continuous 30-minute operations. We leave the exploration of power duration for longer-term operations as future work. We added rubber tape to the foot nodes to improve friction and ensured that the robot had enough pushing force to move forward. For the sim-to-real transfer of the object tracking and loco-manipulation policy, the challenge on ToF cameras primarily lies in the noise from the real sensor reading, the sensor reading delay, and the sensor overheating. To simulate real-world randomness or measurement error during training, we applied $20\%$ random dropout and Gaussian noise following a normal distribution with zero mean and standard deviation of $0.005$ to each measured point. Modeling the thermal behavior of the 20 ToF sensors is beyond the scope of this work and is left for future investigation. We believe that future sensor advancements and selections can largely mitigate the thermal issues.
% Due to the modular design and the symmetry of the spherical structure, system identification conducted on a few legs was generalized well to all. 
% Our sim-to-real transfer efforts were rather standard and minimal.
% \textcolor{blue}{}

\subsection*{Statistical Analysis} Quantitative results consist of deterministic dynamic-isotropy computations, large-scale simulated rollouts, and small-sample physical-robot trials. Dynamic isotropy scores and attainable-acceleration clouds (Figs.~\ref{fig:isotropy_robots}--\ref{fig:argus_sim_isotropy} and Figs.~S\ref{fig:isotropy_trajectory}--S\ref{fig:isotropy_redundancy}) were computed by sampling 2,048 uniformly distributed directions on the unit sphere for each morphology and pose. 

The redundancy analysis in Fig.~S\ref{fig:isotropy_redundancy} reports the relative distance error over 10 equal-width isotropy bins across 1,536 randomized morphologies (512 each for the 12-, 20-, and 32-leg variants), summarized as mean $\pm$ standard deviation within each bin. Within each isotropy bin, pairwise Mann-Whitney U tests were conducted on the relative distance error. All pairwise comparisons within each bin were corrected for multiple comparisons using the Bonferroni method. Significance thresholds were set at * $p < 0.05$, ** $p < 0.01$, and *** $p < 0.001$. Results are presented as box plots showing the median, interquartile range (IQR), and 1.5$\times$IQR whiskers. The sample sizes ($n$) are reported for each group in the figure.

Simulated policies were evaluated in Isaac Gym under randomized initial conditions and the domain randomization (Table~S\ref{table:randomization}), with 8192 trials for baseline locomotion and locomotion under leg failure, self-stabilization, payload carrying, discrete-terrain traversal; and 4096 trials for wall climbing; and 7168 trials for object tracking and object pushing. Unless noted otherwise, shaded regions and error bars denote the 95\% confidence interval, and success rates are reported as the fraction of trials meeting the task-specific criterion defined in each task description above. 

Physical-robot outcomes are reported as trial counts on a single 20-leg prototype: 15 of 18 trials for discrete-terrain traversal, 14 of 38 trials for object tracking, and 13 of 33 trials for object pushing. Data aggregation, statistical summaries, and plotting were performed in Python using NumPy, pandas, and Matplotlib.

%%%%%%%%%%%%%%%% SUPPLEMENT LIST %%%%%%%%%%%%%%%

% List the contents of your Supplementary Materials, including the numbers of any
% supplementary figures, tables, external data files etc. and any references that are
% cited only in the supplement. In this example, refs. 7-8 are cited only in the supplement.
% Fill out your numbers accordingly and delete any lines that aren't applicable.
\section*{Supplementary materials}
Supplementary Methods \\
Figs. S1 to S10\\
Tables S1 to S4\\
Movie 1 \\
Movie S1 to S10
%%%%%%%%%%%%%%%% REFERENCES %%%%%%%%%%%%%%%

\clearpage % Clear all remaining figures and tables then start a new page

% The list of references goes after the main text and before the acknowledgements
% When preparing an initial submission, we recommend you use BibTeX, like this:
%
\bibliography{Argus} % for a file named science_template.bib

@inproceedings{su2024leveraging,
  title={Leveraging symmetry in rl-based legged locomotion control},
  author={Su, Zhi and Huang, Xiaoyu and Ordo{\~n}ez-Apraez, Daniel and Li, Yunfei and Li, Zhongyu and Liao, Qiayuan and Turrisi, Giulio and Pontil, Massimiliano and Semini, Claudio and Wu, Yi},
  booktitle={2024 IEEE/RSJ International Conference on Intelligent Robots and Systems (IROS)},
  pages={6899--6906},
  year={2024},
  organization={IEEE}
}

@article{salisbury1982articulated,
  title={Articulated hands: Force control and kinematic issues},
  author={Salisbury, J Kenneth and Craig, John J},
  journal={Int. J. Robot. Res.},
  volume={1},
  number={1},
  pages={4--17},
  year={1982},
  publisher={Sage Publications Sage UK: London, England}
}

@inproceedings{ma1993optimum,
  title={Optimum design of manipulators under dynamic isotropy conditions},
  author={Ma, Ou and Angeles, Jorge},
  booktitle={[1993] Proceedings IEEE International Conference on Robotics and Automation},
  pages={470--475},
  year={1993},
  organization={IEEE}
}

@inproceedings{ma1990concept,
  title={The concept of dynamic isotropy and its applications to inverse kinematics and trajectory planning},
  author={Ma, Ou and Angeles, Jorge},
  booktitle={Proceedings., IEEE International Conference on Robotics and Automation},
  pages={481--486},
  year={1990},
  organization={IEEE}
}

@article{klein1987dexterity,
  title={Dexterity measures for the design and control of kinematically redundant manipulators},
  author={Klein, Charles A and Blaho, Bruce E},
  journal={Int. J. Robot. Res.},
  volume={6},
  number={2},
  pages={72--83},
  year={1987},
  publisher={Sage Publications Sage UK: London, England}
}

@article{klein1991spatial,
  title={Spatial robotic isotropy},
  author={Klein, Charles A and Miklos, Todd A},
  journal={Int. J. Robot. Res.},
  volume={10},
  number={4},
  pages={426--437},
  year={1991},
  publisher={Sage Publications Sage CA: Thousand Oaks, CA}
}

@article{yoshikawa1985manipulability,
  title={Manipulability of robotic mechanisms},
  author={Yoshikawa, Tsuneo},
  journal={Int. J. Robot. Res.},
  volume={4},
  number={2},
  pages={3--9},
  year={1985},
  publisher={Sage Publications Sage CA: Thousand Oaks, CA}
}

@article{yoshikawa1985dynamic,
  title={Dynamic manipulability of robot manipulators},
  author={Yoshikawa, Tsuneo},
  journal={Trans. Soc. Instrum. Control Eng.},
  volume={21},
  number={9},
  pages={970--975},
  year={1985},
  publisher={The Society of Instrument and Control Engineers}
}

@article{kaelbling1998planning,
  title={Planning and acting in partially observable stochastic domains},
  author={Kaelbling, Leslie Pack and Littman, Michael L and Cassandra, Anthony R},
  journal={Artif. Intell.},
  volume={101},
  number={1-2},
  pages={99--134},
  year={1998},
  publisher={Elsevier}
}

@book{capek2004rur,
  title={RUR (Rossum's universal robots)},
  author={Capek, Karel},
  year={2004},
  publisher={Penguin}
}

@article{glasser1992energies,
  title={Energies and spacings of point charges on a sphere},
  author={Glasser, Leslie and Every, AG},
  journal={J. Phys. A},
  volume={25},
  number={9},
  pages={2473},
  year={1992},
  publisher={IOP Publishing}
}

@article{pfeifer2007self,
  title={Self-organization, embodiment, and biologically inspired robotics},
  author={Pfeifer, Rolf and Lungarella, Max and Iida, Fumiya},
  journal={Science},
  volume={318},
  number={5853},
  pages={1088--1093},
  year={2007},
  publisher={American Association for the Advancement of Science}
}

@article{trivedi2008soft,
author = {Deepak Trivedi and Christopher D. Rahn and William M. Kier and Ian D. Walker},
title = {Soft robotics: Biological inspiration, state of the art, and future research},
journal = {Appl. Bionics Biomech.},
volume = {5},
number = {3},
pages = {99--117},
year = {2008},
publisher = {Taylor \& Francis},
doi = {10.1080/11762320802557865},
}

@incollection{iida2016biologically,
  title={Biologically inspired robotics},
  author={Iida, Fumiya and Ijspeert, Auke Jan},
  booktitle={Springer Handbook of Robotics},
  pages={2015--2034},
  year={2016},
  publisher={Springer}
}

@article{bar2003biologically,
  title={Biologically inspired intelligent robots},
  author={Bar-Cohen, Yoseph and Breazeal, Cynthia},
  journal={Smart Structures and Materials 2003: Electroactive Polymer Actuators and Devices (EAPAD)},
  volume={5051},
  pages={14--20},
  year={2003},
  publisher={SPIE}
}

@article{lepora2013state,
  title={The state of the art in biomimetics},
  author={Lepora, Nathan F and Verschure, Paul and Prescott, Tony J},
  journal={Bioinspir. Biomim.},
  volume={8},
  number={1},
  pages={013001},
  year={2013},
  publisher={IOP Publishing}
}

@inproceedings{dong2023symmetry,
  title={Symmetry-aware robot design with structured subgroups},
  author={Dong, Heng and Zhang, Junyu and Wang, Tonghan and Zhang, Chongjie},
  booktitle={International Conference on Machine Learning},
  pages={8334--8355},
  year={2023},
  organization={PMLR}
}

@article{ghaffari2022progress,
  title={Progress in symmetry preserving robot perception and control through geometry and learning},
  author={Ghaffari, Maani and Zhang, Ray and Zhu, Minghan and Lin, Chien Erh and Lin, Tzu-Yuan and Teng, Sangli and Li, Tingjun and Liu, Tianyi and Song, Jingwei},
  journal={Front. Robot. AI},
  volume={9},
  pages={969380},
  year={2022},
  publisher={Frontiers Media SA}
}

@inproceedings{kim2007systematic,
  title={Systematic isotropy analysis of a mobile robot with three active caster wheels},
  author={Kim, Sungbok and Jeong, Ilhwa and Lee, Sanghyup},
  booktitle={International Conference on Intelligent Computing},
  pages={587--597},
  year={2007},
  organization={Springer}
}

@article{ocklenburg2022symmetry,
  title={Symmetry and asymmetry in biological structures},
  author={Ocklenburg, Sebastian and Mundorf, Annakarina},
  journal={Proc. Natl. Acad. Sci. U.S.A.},
  volume={119},
  number={28},
  pages={e2204881119},
  year={2022},
  publisher={National Academy of Sciences}
}

@article{enquist1994symmetry,
  title={Symmetry, beauty and evolution},
  author={Enquist, Magnus and Arak, Anthony},
  journal={Nature},
  volume={372},
  number={6502},
  pages={169--172},
  year={1994},
  publisher={Nature Publishing Group UK London}
}

@article{raibert2008bigdog,
  title={Bigdog, the rough-terrain quadruped robot},
  author={Raibert, Marc and Blankespoor, Kevin and Nelson, Gabriel and Playter, Rob},
  journal={IFAC Proc. Vol.},
  volume={41},
  number={2},
  pages={10822--10825},
  year={2008},
  publisher={Elsevier}
}

@article{mattar2013survey,
  title={A survey of bio-inspired robotics hands implementation: New directions in dexterous manipulation},
  author={Mattar, Ebrahim},
  journal={Robot. Auton. Syst.},
  volume={61},
  number={5},
  pages={517--544},
  year={2013},
  publisher={Elsevier}
}

@inproceedings{ma2011dexterity,
  title={On dexterity and dexterous manipulation},
  author={Ma, Raymond R and Dollar, Aaron M},
  booktitle={2011 15th International Conference on Advanced Robotics (ICAR)},
  pages={1--7},
  year={2011},
  organization={IEEE}
}

@article{sun2022recent,
  title={Recent progress in modeling and control of bio-inspired fish robots},
  author={Sun, Boai and Li, Weikun and Wang, Zhangyuan and Zhu, Yunpeng and He, Qu and Guan, Xinyan and Dai, Guangmin and Yuan, Dehan and Li, Ang and Cui, Weicheng and Fan, Dixia},
  journal={J. Mar. Sci. Eng.},
  volume={10},
  number={6},
  pages={773},
  year={2022},
  publisher={MDPI}
}

@inproceedings{xia2025duke,
  title={The Duke Humanoid: Design and Control for Energy-Efficient Bipedal Locomotion Using Passive Dynamics},
  author={Xia, Boxi and Li, Bokuan and Lee, Jacob and Scutari, Michael and Chen, Boyuan},
  booktitle={2025 IEEE/RSJ International Conference on Intelligent Robots and Systems (IROS)},
  pages={6579--6586},
  year={2025},
  organization={IEEE}
}

@inproceedings{liao2025berkeley,
  title={Berkeley humanoid: A research platform for learning-based control},
  author={Liao, Qiayuan and Zhang, Bike and Huang, Xuanyu and Huang, Xiaoyu and Li, Zhongyu and Sreenath, Koushil},
  booktitle={2025 IEEE International Conference on Robotics and Automation (ICRA)},
  pages={2897--2904},
  year={2025},
  organization={IEEE}
}

@article{metta2010icub,
  title={The iCub humanoid robot: An open-systems platform for research in cognitive development},
  author={Metta, Giorgio and Natale, Lorenzo and Nori, Francesco and Sandini, Giulio and Vernon, David and Fadiga, Luciano and Von Hofsten, Claes and Rosander, Kerstin and Lopes, Manuel and Santos-Victor, Jos{\'e} and Bernardino, Alexandre and Montesano, Luis},
  journal={Neural Netw.},
  volume={23},
  number={8-9},
  pages={1125--1134},
  year={2010},
  publisher={Elsevier}
}

@inproceedings{saloutos2023design,
  title={Design and development of the mit humanoid: A dynamic and robust research platform},
  author={SaLoutos, Andrew and Stanger-Jones, Elijah and Ding, Yanran and Chignoli, Matthew and Kim, Sangbae},
  booktitle={2023 IEEE-RAS 22nd International Conference on Humanoid Robots (Humanoids)},
  pages={1--8},
  year={2023},
  organization={IEEE}
}

@inproceedings{nozaki2017Shape,
	title = {Shape changing locomotion by spiny multipedal robot},
	doi = {10.1109/ROBIO.2017.8324739},
	urldate = {2025-02-21},
	booktitle = {2017 {IEEE} {International} {Conference} on {Robotics} and {Biomimetics} ({ROBIO})},
	author = {Nozaki, Hiroki and Niiyama, Ryuma and Yonezawa, Takuro and Nakazawa, Jin},
	month = dec,
	year = {2017},
	pages = {2162--2166},
}

@inproceedings{vespignani2018design,
  title={Design of SUPERball v2, a Compliant Tensegrity Robot for Absorbing Large Impacts},
  author={Vespignani, Massimo and Friesen, Jeffrey M. and SunSpiral, Vytas and Bruce, Jonathan},
  booktitle={2018 IEEE/RSJ International Conference on Intelligent Robots and Systems (IROS)},
  pages={2865--2871},
  year={2018},
  organization={IEEE},
  doi={10.1109/IROS.2018.8594233}
}

@article{zheng2021robustness,
  title={Robustness evaluation for rolling gaits of a six-strut tensegrity robot},
  author={Zheng, Yanfeng and Li, Yi and Lu, Yipeng and Wang, Meijia and Xu, Xian and Zhou, Chunlin and Luo, Yaozhi},
  journal={Int. J. Adv. Robot. Syst.},
  volume={18},
  number={1},
  pages={1--11},
  year={2021},
  publisher={SAGE Publications},
  doi={10.1177/1729881421993638}
}

@inproceedings{nozaki2018Continuous,
	title = {Continuous {Shape} {Changing} {Locomotion} of 32-legged {Spherical} {Robot}},
	doi = {10.1109/IROS.2018.8593791},
	urldate = {2025-02-20},
	booktitle = {2018 {IEEE}/{RSJ} {International} {Conference} on {Intelligent} {Robots} and {Systems} ({IROS})},
	author = {Nozaki, Hiroki and Kujirai, Yusei and Niiyama, Ryuma and Kawahara, Yoshihiro and Yonezawa, Takuro and Nakazawa, Jin},
	month = oct,
	year = {2018},
	pages = {2721--2726},
}

@inproceedings{yang2023bionic,
	title = {Bionic {Multi}-legged {Robot} {Based} on {End}-to-end {Artificial} {Neural} {Network} {Control}},
	doi = {10.1109/CBS55922.2023.10115331},
	urldate = {2025-02-20},
	booktitle = {2022 {IEEE} {International} {Conference} on {Cyborg} and {Bionic} {Systems} ({CBS})},
	author = {Yang, Dun and Liu, Yunfei and Ding, Fei and Yu, Yang},
	month = mar,
	year = {2023},
	pages = {104--109},
}

@inproceedings{wang2023robot,
  title={On-Robot Learning With Equivariant Models},
  author={Wang, Dian and Jia, Mingxi and Zhu, Xupeng and Walters, Robin and Platt, Robert},
  booktitle={Conference on Robot Learning},
  pages={1345--1354},
  year={2023},
  organization={PMLR}
}

@inproceedings{yan2024learning,
  title={Learning continuous control with geometric regularity from robot intrinsic symmetry},
  author={Yan, Shengchao and Zhang, Baohe and Zhang, Yuan and Boedecker, Joschka and Burgard, Wolfram},
  booktitle={2024 IEEE International Conference on Robotics and Automation (ICRA)},
  pages={49--55},
  year={2024},
  organization={IEEE}
}

@inproceedings{zhu2022sample,
  title={Sample Efficient Grasp Learning Using Equivariant Models},
  author={Zhu, Xupeng and Wang, Dian and Biza, Ondrej and Su, Guanang and Walters, Robin and Platt, Robert},
  booktitle={Robotics: Science and Systems},
  year={2022}
}

@inproceedings{mittal2024symmetry,
  title={Symmetry considerations for learning task symmetric robot policies},
  author={Mittal, Mayank and Rudin, Nikita and Klemm, Victor and Allshire, Arthur and Hutter, Marco},
  booktitle={2024 IEEE International Conference on Robotics and Automation (ICRA)},
  pages={7433--7439},
  year={2024},
  organization={IEEE}
}

@misc{schulman2017proximal,
  title={Proximal policy optimization algorithms},
  author={Schulman, John and Wolski, Filip and Dhariwal, Prafulla and Radford, Alec and Klimov, Oleg},
  year={2017},
  howpublished={\url{https://arxiv.org/abs/1707.06347}}
}

@misc{makoviychuk2021isaac,
  title={Isaac gym: High performance gpu-based physics simulation for robot learning},
  author={Makoviychuk, Viktor and Wawrzyniak, Lukasz and Guo, Yunrong and Lu, Michelle and Storey, Kier and Macklin, Miles and Hoeller, David and Rudin, Nikita and Allshire, Arthur and Handa, Ankur and State, Gavriel},
  year={2021},
  howpublished={\url{https://arxiv.org/abs/2108.10470}}
}

@article{shin2024fast,
  title={Fast ground-to-air transition with avian-inspired multifunctional legs},
  author={Shin, Won Dong and Phan, Hoang-Vu and Daley, Monica A and Ijspeert, Auke J and Floreano, Dario},
  journal={Nature},
  volume={636},
  number={8041},
  pages={86--91},
  year={2024},
  publisher={Nature Publishing Group UK London}
}

@article{langowski2020soft,
  title={In the soft grip of nature},
  author={Langowski, JKA and Sharma, Preeti and Shoushtari, A Leylavi},
  journal={Sci. Robot.},
  volume={5},
  number={49},
  pages={eabd9120},
  year={2020},
  publisher={American Association for the Advancement of Science}
}

@article{jeong_spikebot_2024,
	title = {Spikebot: {A} {Multigait} {Tensegrity} {Robot} with {Linearly} {Extending} {Struts}},
	volume = {11},
	issn = {2169-5172},
	shorttitle = {Spikebot},
	doi = {10.1089/soro.2023.0030},
	number = {2},
	urldate = {2025-02-20},
	journal = {Soft Robot.},
	author = {Jeong, Jinwook and Kim, Injoong and Choi, Yunyeong and Lim, Seonghyeon and Kim, Seungkyu and Kang, Hyeongwoo and Shah, Dylan and Baines, Robert and Booth, Joran W. and Kramer-Bottiglio, Rebecca and Kim, Sang Yup},
	month = apr,
	year = {2024},
	pages = {207--217},
}

@inproceedings{gheorghe_rolling_2010,
	title = {Rolling robot with radial extending legs},
	doi = {10.1109/ISRCS.2010.5603951},
	urldate = {2025-02-20},
	booktitle = {2010 3rd {International} {Symposium} on {Resilient} {Control} {Systems}},
	author = {Gheorghe, Viorel and Alexandrescu, Nicolae and Duminica, Despina and Cartal, Laurentiu Adrian},
	month = aug,
	year = {2010},
	pages = {107--112},
}

@article{xu2024physics,
	title = {A {Physics}-{Driven} {Closed}-{Loop} {Motion} {Planning} {Method} for {Spherical} {Multi}-{Expandable}-{Limb} {Robots}},
	volume = {71},
	issn = {1557-9948},
	doi = {10.1109/TIE.2024.3390725},
	number = {12},
	urldate = {2025-02-20},
	journal = {IEEE Trans. Ind. Electron.},
	author = {Xu, Fengde and Zhao, Xudong and Yue, Ming},
	month = dec,
	year = {2024},
	pages = {16087--16097},
}

@inproceedings{yang_general_2023,
	title = {A {General} {Locomotion} {Approach} for a {Novel} {Multi}-legged {Spherical} {Robot}},
	doi = {10.1109/ICRA48891.2023.10160881},
	urldate = {2025-02-20},
	booktitle = {2023 {IEEE} {International} {Conference} on {Robotics} and {Automation} ({ICRA})},
	author = {Yang, Dun and Liu, Yunfei and Yu, Yang},
	month = may,
	year = {2023},
	pages = {10146--10152},
}

@article{kim_rolling_2020,
	title = {Rolling {Locomotion} of {Cable}-{Driven} {Soft} {Spherical} {Tensegrity} {Robots}},
	volume = {7},
	issn = {2169-5172},
	doi = {10.1089/soro.2019.0056},
	number = {3},
	urldate = {2025-07-31},
	journal = {Soft Robot.},
	author = {Kim, Kyunam and Agogino, Adrian K. and Agogino, Alice M.},
	month = jun,
	year = {2020},
	pmid = {32031916},
	pmcid = {PMC7301328},
	pages = {346--361},
}

@article{liu_design_2020,
	title = {Design and analysis of a deployable tetrahedron-based mobile robot constructed by {Sarrus} linkages},
	volume = {152},
	issn = {0094114X},
	doi = {10.1016/j.mechmachtheory.2020.103964},
	language = {en},
	urldate = {2025-02-21},
	journal = {Mech. Mach. Theory},
	author = {Liu, Ran and Yao, Yan-an and Li, Yezhuo},
	month = oct,
	year = {2020},
	pages = {103964},
}

@article{surovik_adaptive_2021,
	title = {Adaptive tensegrity locomotion: {Controlling} a compliant icosahedron with symmetry-reduced reinforcement learning},
	volume = {40},
	issn = {0278-3649},
	shorttitle = {Adaptive tensegrity locomotion},
	doi = {10.1177/0278364919859443},
	number = {1},
	urldate = {2025-02-21},
	journal = {Int. J. Robot. Res.},
	author = {Surovik, David and Wang, Kun and Vespignani, Massimo and Bruce, Jonathan and Bekris, Kostas E},
	month = jan,
	year = {2021},
	pages = {375--396}
}

@inproceedings{pai1994platonic,
  title={Platonic beasts: a new family of multilimbed robots},
  author={Pai, Dinesh K and Barman, Roderick A and Ralph, Scott K},
  booktitle={Proceedings of the 1994 IEEE International Conference on Robotics and Automation},
  pages={1019--1025},
  year={1994},
  organization={IEEE}
}

@inproceedings{qi2017pointnet,
  title={Pointnet: Deep learning on point sets for 3d classification and segmentation},
  author={Qi, Charles R and Su, Hao and Mo, Kaichun and Guibas, Leonidas J},
  booktitle={Proceedings of the IEEE conference on computer vision and pattern recognition},
  pages={652--660},
  year={2017}
}

@article{paul2006design,
  title={Design and control of tensegrity robots for locomotion},
  author={Paul, Chandana and Valero-Cuevas, Francisco J and Lipson, Hod},
  journal={IEEE Trans. Robot.},
  volume={22},
  number={5},
  pages={944--957},
  year={2006},
  publisher={Ieee}
}

@inproceedings{bledt2018cheetah,
  title={Mit cheetah 3: Design and control of a robust, dynamic quadruped robot},
  author={Bledt, Gerardo and Powell, Matthew J and Katz, Benjamin and Di Carlo, Jared and Wensing, Patrick M and Kim, Sangbae},
  booktitle={2018 IEEE/RSJ International Conference on Intelligent Robots and Systems (IROS)},
  pages={2245--2252},
  year={2018},
  organization={IEEE}
}

@inproceedings{hutter2016anymal,
  title={Anymal-a highly mobile and dynamic quadrupedal robot},
  author={Hutter, Marco and Gehring, Christian and Jud, Dominic and Lauber, Andreas and Bellicoso, C Dario and Tsounis, Vassilios and Hwangbo, Jemin and Bodie, Karen and Fankhauser, Peter and Bloesch, Michael and Diethelm, Remo and Bachmann, Samuel and Melzer, Amir and Hoepflinger, Mark},
  booktitle={2016 IEEE/RSJ international conference on intelligent robots and systems (IROS)},
  pages={38--44},
  year={2016},
  organization={IEEE}
}

@inproceedings{wang2022equivariant,
  title={Equivariant $ q $ learning in spatial action spaces},
  author={Wang, Dian and Walters, Robin and Zhu, Xupeng and Platt, Robert},
  booktitle={Conference on Robot Learning},
  pages={1713--1723},
  year={2022},
  organization={PMLR}
}

@article{frazzoli2005maneuver,
  title={Maneuver-based motion planning for nonlinear systems with symmetries},
  author={Frazzoli, Emilio and Dahleh, Munther A and Feron, Eric},
  journal={IEEE Trans. Robot.},
  volume={21},
  number={6},
  pages={1077--1091},
  year={2005},
  publisher={IEEE}
}

@article{apraez2025morphological,
  title={Morphological symmetries in robotics},
  author={Apraez, Daniel Ordonez and Turrisi, Giulio and Kostic, Vladimir and Martin, Mario and Agudo, Antonio and Moreno-Noguer, Francesc and Pontil, Massimiliano and Semini, Claudio and Mastalli, Carlos},
  journal={Int. J. Robot. Res.},
  volume={44},
  pages={1743--1766},
  year={2024}
}

@article{chang2024bird,
  title={Bird-inspired reflexive morphing enables rudderless flight},
  author={Chang, Eric and Chin, Diana D and Lentink, David},
  journal={Sci. Robot.},
  volume={9},
  number={96},
  pages={eado4535},
  year={2024},
  publisher={American Association for the Advancement of Science}
}

@article{li2021self,
  title={Self-powered soft robot in the Mariana Trench},
  author={Li, Guorui and Chen, Xiangping and Zhou, Fanghao and Liang, Yiming and Xiao, Youhua and Cao, Xunuo and Zhang, Zhen and Zhang, Mingqi and Wu, Baosheng and Yin, Shunyu and Xu, Yi and Fan, Hongbo and Chen, Zheng and Song, Wei and Yang, Wenjing and Pan, Binbin and Hou, Jiaoyi and Zou, Weifeng and He, Shunping and Yang, Xuxu and Mao, Guoyong and Jia, Zheng and Zhou, Haofei and Li, Tiefeng and Qu, Shaoxing and Xu, Zhongbin and Huang, Zhilong and Luo, Yingwu and Xie, Tao and Gu, Jason and Zhu, Shiqiang and Yang, Wei},
  journal={Nature},
  volume={591},
  number={7848},
  pages={66--71},
  year={2021},
  publisher={Nature Publishing Group UK London}
}

@article{baines2022multi,
  title={Multi-environment robotic transitions through adaptive morphogenesis},
  author={Baines, Robert and Patiballa, Sree Kalyan and Booth, Joran and Ramirez, Luis and Sipple, Thomas and Garcia, Andonny and Fish, Frank and Kramer-Bottiglio, Rebecca},
  journal={Nature},
  volume={610},
  number={7931},
  pages={283--289},
  year={2022},
  publisher={Nature Publishing Group UK London}
}

@article{liu2025lcrbot,
  title={LCRBot: Load-Carrying Rolling Robot Based on Truncated Hexahedral Tensegrity},
  author={Liu, Jilei and Xu, Zhiyin and Lu, Jinyu and Gu, Xun and Wu, Jiarong},
  journal={J. Field Robot.},
  volume={42},
  number={6},
  pages={2454--2468},
  year={2025}
}

@inproceedings{chen2024learning,
  title={Learning Differentiable Tensegrity Dynamics using Graph Neural Networks},
  author={Chen, Nelson and Wang, Kun and Johnson III, William R. and Kramer-Bottiglio, Rebecca and Bekris, Kostas and Aanjaneya, Mridul},
  booktitle={Proceedings of the Conference on Robot Learning (CoRL)},
  year={2024},
}

@misc{dasgupta_spiderbot_deeprl,
  author       = {Dasgupta, Arijit},
  title        = {SpiderBot\_DeepRL: A Custom-Designed Spider Robot Trained to Walk Using Deep Reinforcement Learning in a PyBullet Simulation},
  year         = {2026},
  howpublished = {\url{https://github.com/arijit-dasgupta/SpiderBot_DeepRL}},
  note         = {GitHub repository. Accessed: 2026-02-23}
}

@misc{awesome_robot_descriptions,
  author       = {robot-descriptions},
  title        = {awesome-robot-descriptions},
  howpublished = {GitHub repository},
  year         = {2026},
  url          = {https://github.com/robot-descriptions/awesome-robot-descriptions},
  note         = {Accessed: 2026-02-23}
}
\bibliographystyle{sciencemag}

% After the paper has completed peer review and been revised ready for acceptance,
% you should comment out the lines above and copy-paste the contents of your .bbl
% file here instead. This will help ensure that our conversion software works correctly.
% Remember to re-run BibTeX first - check the timestamp!
%
% Example of the first three entries copy-pasted from science_template.bbl:
%
%\begin{thebibliography}{1}
%
%\bibitem{example}
%A.~N. {Author}, An example reference. \emph{Journal of Improbable Research}
%  \textbf{1}, 67 (2020).
%
%\bibitem{example2}
%F.~M. {Surname}, S.~{Author}, A second example. \emph{Interesting Research
%  Letters} \textbf{32}, 897 (2019).
%
%\bibitem{example_preprint}
%P.~{One}, P.~{Two}, P.~{Three}, {An unpublished preprint}. \emph{preprint}
%  (2021), arXiv:2101.12345.
%
%\end{thebibliography}

%%%%%%%%%%%%%%%% ACKNOWLEDGEMENTS %%%%%%%%%%%%%%%

\noindent \textbf{Acknowledgments}: The authors would like to thank Edwin Ma, Jacob Lee, and Max Li for their help in 3D printing. We thank Sam Moore for the advice on the implementation of ray casting and Yinsen Jia for the reinforcement learning implementation. \textbf{Funding}: This work is supported by DARPA FoundSci program under award HR00112490372, DARPA TIAMAT program under award HR00112490419, ARO under award W911NF2410405, ARL STRONG program under awards W911NF2320182, W911NF2220113, and W911NF242021. \textbf{Author contributions}: B.C., J.L., and B.X. conceived and designed the research. J.L. and B.X. designed simulations and performed physical experiments. All authors analyzed data and wrote the manuscript. \textbf{Competing interests}: Duke University has filed patent rights for the technology associated with this manuscript. For further license rights, including using the patent rights for commercial purposes, please contact Duke's Office for Translation and Commercialization (otcquestions@duke.edu) and reference OTC DU8860PROV. \textbf{Data, code and materials availability}:
% The code is available at https://doi.org/10.5061/dryad.3j9kd520k. 
The code will be available once the paper is accepted.
The high-resolution version of all the supplementary videos can be accessed at  https://figshare.com/s/0e117c0158f70c39c6b8. Materials were commercially available.

%%%%%%%%%%%%%%%% END OF MAIN TEXT %%%%%%%%%%%%%%%

\newpage

%%%%%%%%%%%%%%%% START OF SUPPLEMENT %%%%%%%%%%%%%%%

%%%%%%%%%%%%%%%% SUPPLEMENT TITLE PAGE %%%%%%%%%%%%%%%

\begin{center}
    \section*{Supplementary Materials for\\ \scititle}

\author{
	Jiaxun Liu$^{1\dagger}$, Boxi Xia$^{1\dagger}$, Boyuan Chen$^{1, 2, 3\ast}$ \\
    \normalsize{$^{1}$Department of Mechanical Engineering and Materials Science, Duke University}\\
    \normalsize{$^{2}$Department of Electrical and Computer Engineering, Duke University}\\
    \normalsize{$^{3}$Department of Computer Science, Duke University}\\
    \normalsize{$^\ast$To whom correspondence should be addressed; E-mail: boyuan.chen@duke.edu.}\\
    \normalsize{$^\dagger$These authors contributed equally to this work.}
}
\author

\end{center}

% Fill out the numbers for each type of supplementary material,
% and delete any lines that aren't applicable.
% These are just example numbers that don't match the rest of this template.
\subsubsection*{This PDF file includes:}
Supplementary Methods \\
Figures S1 to S10\\
Tables S1 to S4\\
Captions for Movies 1 \\
Captions for Movies S1 to S10

\subsubsection*{Other Supplementary Materials for this manuscript:}
Movie 1 \\
Movies S1 to S10 \\
Data S1
\newpage

% \section*{Dynamic Isotropy Measurement for Various Robots}

\section*{Supplementary Methods}

We present more detailed descriptions and analyses on how our dynamic isotropy framework can generalize to various robots and configurations beyond what we have discussed for Argus variations. All rigid-body legged robots use the same analytical model as Argus, except for the soft tensegrity robot and the drone, which are discussed in the corresponding sections below.

Robot designs closely related to Argus include the Spiny robot\cite{nozaki2017Shape} and Mochibot\cite{nozaki2018Continuous}. The Spiny robot shares the same leg configuration as the 12-legged Argus design. Mochibot places its legs on the 32 vertices of a rhombic triacontahedron. Because these vertices are formed by either five or three rhombi, they are not evenly distributed compared to the optimally arranged 32-legged Argus design. This geometric difference leads to a lower dynamic isotropy score of 0.884 for Mochibot, whereas the 32-legged Argus achieves a score of 0.935.

The tensegrity robot accelerates its center of mass (CoM) through coordinated actuation of its cables and rigid rods. This actuation mechanism reshapes internal force distributions and, through ground contact, generates a net force on the system. For the symmetric 6-bar, 24-cable configuration, the actuation-to-CoM acceleration mapping yields a dynamic isotropy score of 0.867.

For a traditional quadrotor drone, the dynamic isotropy metric captures the non-uniformity arising from its single-directional actuation. Each propeller can only generate maximum thrust upward relative to the center of mass, and therefore naturally leads to zero acceleration in its opposite direction. In contrast, the HAGAMOSphere drone (Hybrid Autonomous Ground / Aerial MObility System, research prototyp, DIC Corporation, Japan) arranges eight propellers symmetrically around a cubic frame, enabling a more symmetric range of linear acceleration directions. As a result, it achieves a higher dynamic isotropy score, comparable to that of the 6-legged Argus.

The robot definitions for the SpiderBot, humanoid, quadruped, and tensegrity robots were obtained from open-source repositories \cite{dasgupta_spiderbot_deeprl, awesome_robot_descriptions} and prior work \cite{chen2024learning}.

\subsection*{Center-of-Mass Linear Acceleration Model for Tensegrity Robots}

We consider tensegrity robots whose $n$ actuator elements are either rigid rods acting as bilateral linear actuators, or cables providing unilateral tensile actuation.
Each actuator element $i$ connects nodes $p_i$ and $q_i$ along the unit direction vector $d_i \in \mathbb{R}^3$:
\begin{equation}
    d_i = \frac{\mathbf{p}_{q_i} - \mathbf{p}_{p_i}}{\|\mathbf{p}_{q_i} - \mathbf{p}_{p_i}\|}.
    \label{eq:elem_dir}
\end{equation}
A rod exerts a controllable axial force $f_i \in [-f_i^{\max},\, +f_i^{\max}]$ (bilateral), whereas a cable exerts a tensile force $f_i \in [0,\, f_i^{\max}]$ (unilateral).
Because all actuator forces are internal, no combination of inputs can accelerate the center of mass (CoM) without external reaction from ground contact. To compute the maximum ever possible acceleration, we omit gravity and all other external forces and assume ideal contact: for any desired acceleration direction, at least one end of each actuator element is in contact with the ground to absorb the reaction force.

For rods, since $f_i$ is bilateral, grounding either endpoint yields the same achievable acceleration set along its axis.
For cables, the choice of grounded endpoint matters because cables can only pull: grounding node $p_i$ produces a net external force of $-f_i\,d_i$ on the system, while grounding node $q_i$ produces $+f_i\,d_i$.
However, the ideal contact assumption permits choosing the endpoint that maximizes the contribution along any query direction $\bm{u}$, i.e.\ $\sigma_i^\star = \operatorname{sign}(\bm{u}^\top d_i)$, which guarantees a non-negative projection and eliminates the unilateral constraint.
Passive (non-actuated) elements are not explicitly modeled in this bound. Thus, the result does not capture constraint coupling and should be interpreted as an upper bound.

\mypara{Geometric acceleration mapping}
When one endpoint of actuator element $i$ is grounded (with optimal endpoint choice for cables), we upper-bound the feasible net external force by assuming that grounding one endpoint allows the actuator to realize an effective external force of magnitude $f_i$ along direction $d_i$.
% the ground absorbs the reaction, leaving a single unbalanced force $f_i\,d_i$.
Summing over all actuator elements and applying Newton's second law to the CoM gives
\begin{equation}
    a_c
    = \frac{1}{m} \sum_{i=1}^{n} f_i\, d_i
    = \boldsymbol{\mathrm{A}}\,\boldsymbol{\mathrm{f}},
    \label{eq:Ageo}
\end{equation}
where $\boldsymbol{\mathrm{f}} = [f_1,\ldots,f_{n}]^\top$ and
\begin{equation}
    \boldsymbol{\mathrm{A}} := \frac{1}{m}
    \begin{bmatrix} d_1 & d_2 & \cdots & d_{n} \end{bmatrix}
    \in \mathbb{R}^{3 \times n}
    \label{eq:Ageo_def}
\end{equation}
is the geometric acceleration mapping.
Each column $\boldsymbol{\mathrm{A}}_i \in \mathbb{R}^3$ represents the CoM acceleration per unit actuator force.
The per-element feasible intervals are $\mathcal{F}_i = [-f_i^{\max},\, f_i^{\max}]$ for rods and $\mathcal{F}_i = [0,\, f_i^{\max}]$ for cables, with the overall feasible set $\mathcal{F} = \prod_{i=1}^{n} \mathcal{F}_i$.

\mypara{Directional maximum acceleration}
Given a query direction $\bm{u} \in \mathbb{S}^2$, the directional sensitivity of actuator $i$ is
\begin{equation}
    c_i(\boldsymbol{\mathrm{u}}) = \boldsymbol{\mathrm{u}}^\top \boldsymbol{\mathrm{A}}_i.
\end{equation}
The maximum achievable acceleration along $\bm{u}$ is
\begin{equation}
    a_{\max}(\boldsymbol{\mathrm{u}})
    = \max_{f\in\mathcal{F}}\; \boldsymbol{\mathrm{u}}^\top \boldsymbol{\mathrm{A}}\, \boldsymbol{\mathrm{f}}
    = \sum_{i=1}^{n}
      \max_{f_i\in\mathcal{F}_i} c_i(\boldsymbol{\mathrm{u}})\, f_i,
    \label{eq:decompose}
\end{equation}
where $\mathcal{F}_i$ is the feasible interval for actuator $i$ and the second equality holds because each $f_i$ is independently bounded, so the joint maximization decomposes into $n$ independent per-element problems.
For a bilateral rod, $\mathcal{F}_i = [-f_i^{\max},\, f_i^{\max}]$ and the sub-problem is solved by $f_i^\star = \operatorname{sign}(c_i(\bm{u}))\, f_i^{\max}$.
For a unilateral cable, $\mathcal{F}_i = [0,\, f_i^{\max}]$; optimal grounding ensures $c_i(\bm{u}) \geq 0$, so the sub-problem is solved by $f_i^\star = f_i^{\max}$.
In both cases, the contribution of actuator $i$ is $|c_i(\bm{u})|\, f_i^{\max}$.
Substituting yields the closed-form result:
\begin{equation}
    a_{\max}(\boldsymbol{\mathrm{u}})
    = \sum_{i=1}^{n}
      \bigl|c_i(\boldsymbol{\mathrm{u}})\bigr|\; f_i^{\max}.
    \label{eq:amax}
\end{equation}
}

\subsection*{Center-of-Mass Linear Acceleration Model for Drone}

The translational dynamics of a rigid multirotor Unpiloted Aerial Vehicle (UAV) are governed by the Newton--Euler equations. 
Let $m$ denote the total mass of the drone, and let $f_i$ denote the thrust magnitude produced by rotor $i$ along its unit direction vector $d_i \in \mathbb{R}^3$, with a total number of $n$ rotors.

To focus on the effect of the rotors on the robot’s acceleration, we analyze the dynamics under quasi-static or low-velocity conditions with gravity compensation. Under these assumptions, the total force applied to the center of mass (CoM) is
\begin{equation}
F_{c} = \sum_{i=1}^{n} f_i d_i .
\end{equation}

The CoM acceleration therefore is
\begin{equation}
\bm{a}_{c} 
= 
\frac{1}{m} \sum_{i=1}^{n} f_i d_i .
\label{eq:com_acc}
\end{equation}
Define the thrust input vector 
\[
\boldsymbol{\mathrm{f}} = [f_1, \dots, f_n]^\top
\]
and the actuation matrix
\begin{equation}
\boldsymbol{\mathrm{A}} = \frac{1}{m}
\begin{bmatrix}
d_1 & d_2 & \cdots & d_n
\end{bmatrix}
\end{equation}
Then the CoM acceleration can be written compactly as
\begin{equation}
\boldsymbol{\mathrm{a}}_{c} = \boldsymbol{\mathrm{A}} \boldsymbol{\mathrm{f}}.
\end{equation}
Each column of $\boldsymbol{\mathrm{A}}$ represents the acceleration contribution per unit thrust of rotor $i$ to the CoM acceleration. 

\mypara{Directional maximum acceleration}
For each unit direction \( \bm{u} \in \mathbb{S}^2 \),the directional sensitivity of actuator \(i\) is
\begin{equation}
c_i(\boldsymbol{\mathrm{u}}) = \boldsymbol{\mathrm{u}}^\top \boldsymbol{\mathrm{A}}_i,
\end{equation}
Assuming independent thrust bounds
\[
\mathcal{F} = \{ \boldsymbol{\mathrm{f}} \mid  0\le f_i \le f_i^{\max} \},
\]
the maximum achievable CoM acceleration magnitude in direction $\bm{u}$ is obtained by saturating each rotor in the sign that maximizes projection onto $\bm{u}$:
\begin{equation}
a_{\max}(\boldsymbol{\mathrm{u}})
= \max_{\boldsymbol{\mathrm{f}} \in \mathcal{F}}
  \boldsymbol{\mathrm{u}}^\top \boldsymbol{\mathrm{A}}\boldsymbol{\mathrm{f}}
= \sum_{i=1}^{n} \big| c_i(\boldsymbol{\mathrm{u}}) \big| f_i^{\max}.
\end{equation}

\subsection*{Dynamic Isotropy from Maximum Acceleration}
For completeness, we restate the dynamic isotropy definition used throughout this work. The following definition applies identically to the legged, tensegrity, and aerial models derived above. We sample a set of directions $\{ \boldsymbol{\mathrm{u}}_k \}_{k=1}^{K}$ uniformly distributed on the unit sphere and compute the directional accelerations $a_{\max}(\boldsymbol{\mathrm{u}}_k)$ of the CoM. The dynamic isotropy is then defined as
\begin{equation}
\upeta
= \frac{a_{\min}}{a_{\max}},
\qquad
a_{\min} = \min_k a_{\max}(\boldsymbol{\mathrm{u}}_k),
\quad
a_{\max} = \max_k a_{\max}(\boldsymbol{\mathrm{u}}_k),
\end{equation}
where $\upeta \in [0,1]$. A perfectly isotropic acceleration capability corresponds to $\upeta = 1$, indicating equal maximum CoM acceleration magnitude in all directions.

\newpage
\begin{extfigure}[!ht] % use extfigure instead!
    \centering\includegraphics[width=0.9\textwidth]{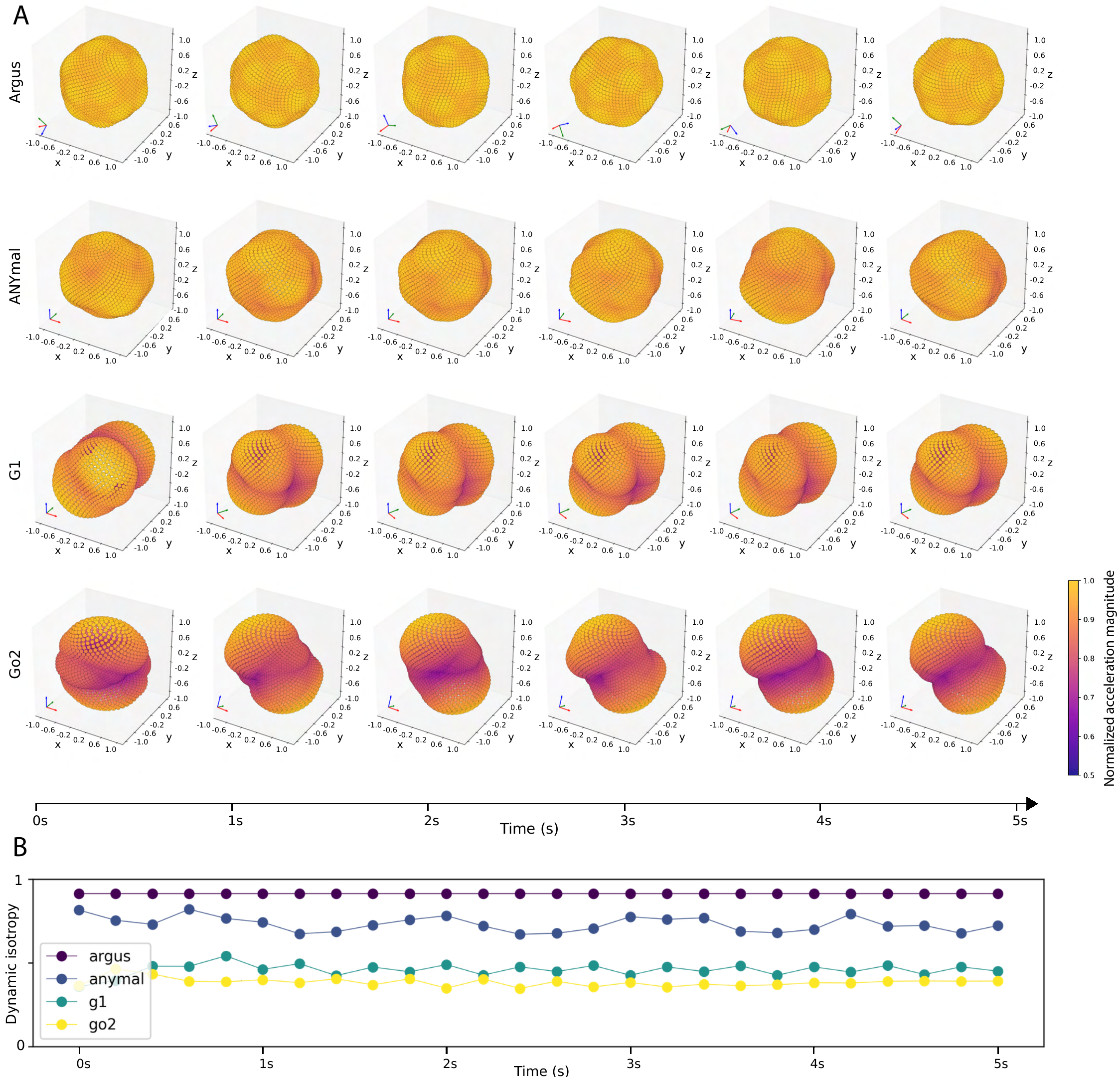}
    \caption{ {\textbf{Dynamic isotropy during forward locomotion across different robot morphologies.} \textbf{(A)}, Attainable acceleration clouds sampled uniformly over 2,048 directions whereas each robot tracks a forward velocity of $0.8\,\mathrm{m/s}$. Argus maintains a nearly spherical acceleration set throughout rolling, whereas ANYmal, G1, and Go2 exhibit strongly anisotropic, posture-dependent acceleration patterns. Although Argus’s acceleration orientation rotates with its body during rolling, the shape of the acceleration cloud remains nearly unchanged, reflecting its high and configuration-invariant dynamic isotropy. \textbf{(B)}, Dynamic isotropy over time for the same trajectories. Argus consistently achieves the highest dynamic isotropy and exhibits minimal fluctuation despite continuous changes in joint configuration and base orientation, whereas quadruped and humanoid robots show lower and more variable dynamic isotropy due to the changes in their joint configurations.}} \label{fig:isotropy_trajectory}
\end{extfigure}
\newpage

\begin{extfigure}[!ht] % use extfigure instead!
    \centering
    \includegraphics[width=1\textwidth]{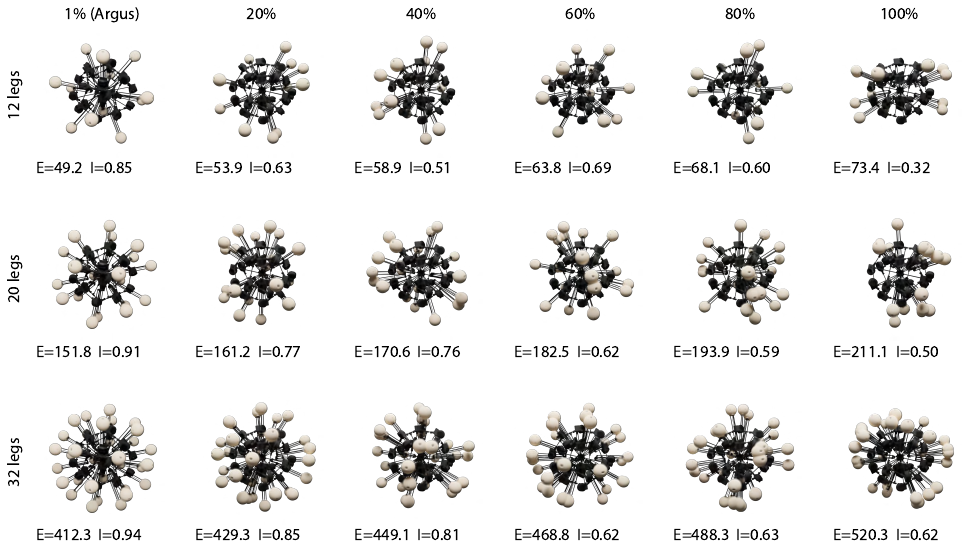}
    \caption{ {\textbf{Argus configurations}: 12, 20, and 32 leg variants are shown on each row, each column shows the variants ranked by the dynamic isotropy (I) and Thomson Energy (E). The first column is the Argus configuration that is most symmetric in terms of dynamic isotropy and Thomson Energy.}}
    \label{fig:dof_vs_energy_isotropy}
\end{extfigure}
\newpage

\begin{extfigure}[!ht] % use extfigure instead!
    \centering
    \includegraphics[width=0.7\textwidth]{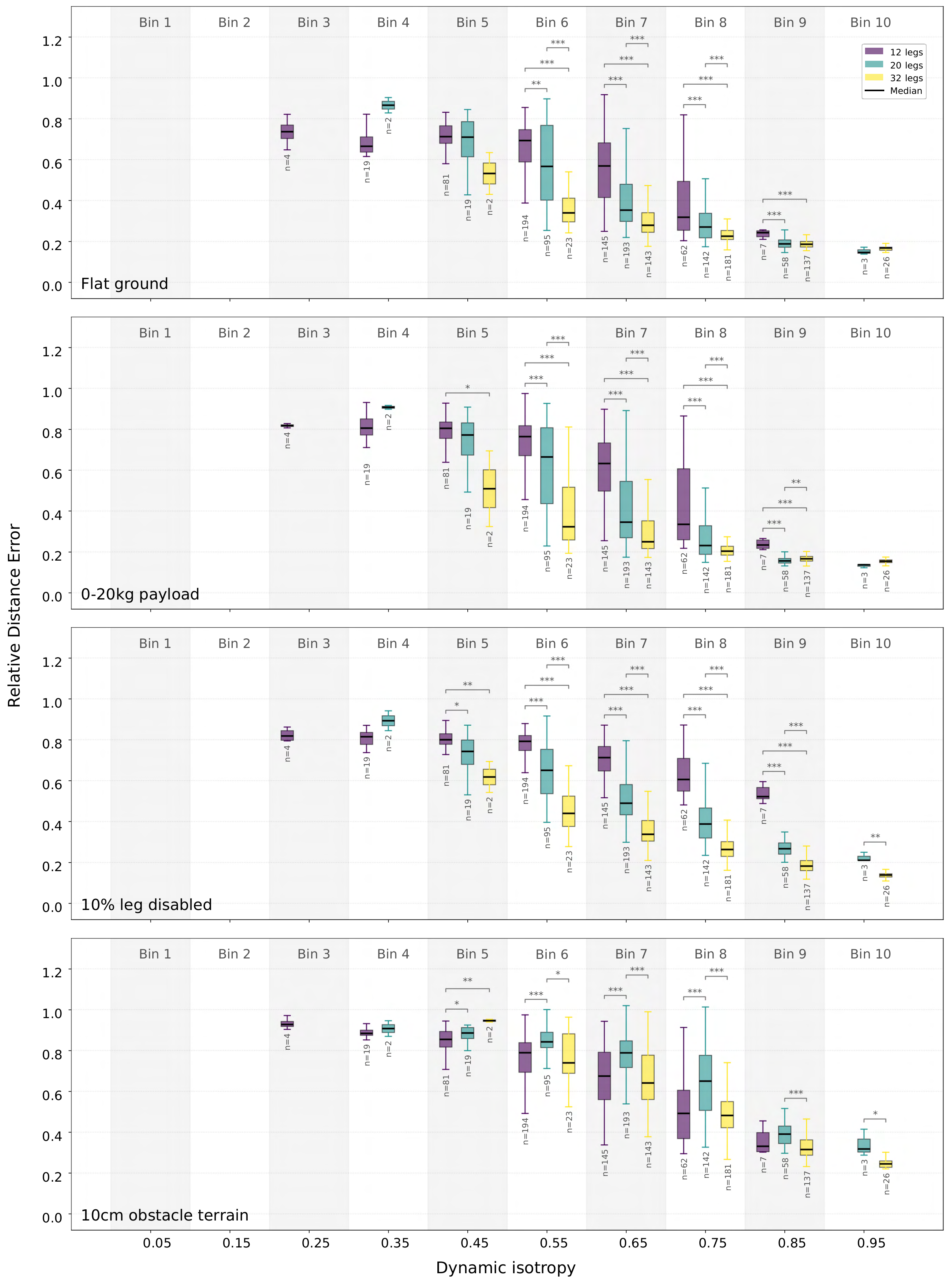}
    \caption{ {\textbf{The redundancy analysis in number of legs.} The mean value and standard deviation of the relative distance error in the large-scale robot learning experiments in simulation in the main text, computed by dividing the isotropy scores into 10 bins. Within each bin, results are reported as mean ± standard deviation across all samples in that bin. The total number of robots is the same for the 12, 20, and 32 leg configurations. However, the number of robots in each bin varies because the designs are randomly sampled, and the dynamic isotropy distribution is centered around the middle. The sample count for each bin is indicated in the figure. This plot supports the analysis of redundancy in the number of legs when the dynamic isotropy values are the same.}}
    \label{fig:isotropy_redundancy}
\end{extfigure}
\newpage

\begin{extfigure}[htbp]
	\centering
    \includegraphics[width=1\textwidth]{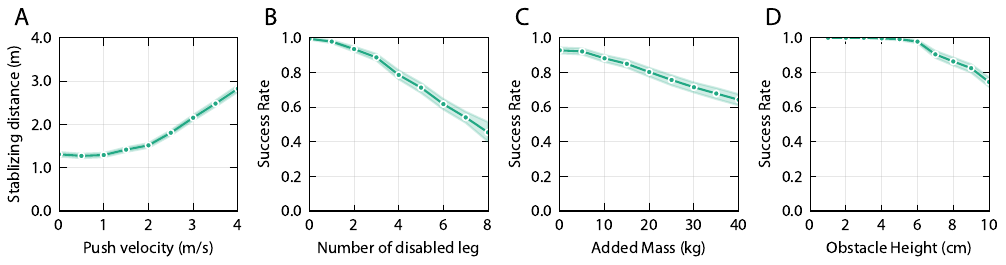}
    \vspace{-10pt}
	\caption{{ \textbf{Simulation evaluations of Argus with 20 legs across four tasks}: \textbf{(A)}, recovery under external pushes, \textbf{(B)}, locomotion with randomly disabled legs, \textbf{(C)}, payload carrying, and \textbf{(D)}, traversal over obstacles. Shaded regions denote the 95\% confidence interval across evaluation trials (8192 samples).
}}
    \label{fig:argus_12_20_32_sim_eval}
\end{extfigure}
\newpage

\begin{extfigure}[!ht]
    \centering
    \includegraphics[width=1\textwidth]{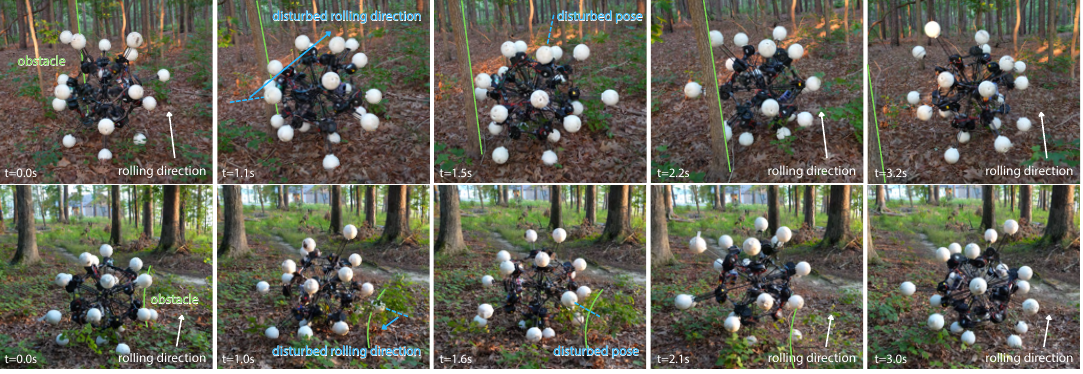}
    \caption{ \textbf{Orientation-invariant locomotion in the wild.} Due to its high isotropy condition, different orientations of the robot exhibit equivalent configurations. When its orientation was passively disturbed by obstacles, Argus continued to track the desired velocity command, remaining unaffected by the change in orientation.}
    \label{fig:extend_invariant_orientation}
\end{extfigure}

\newpage
\begin{extfigure}[!ht]
    \centering
    \includegraphics[width=1\textwidth]{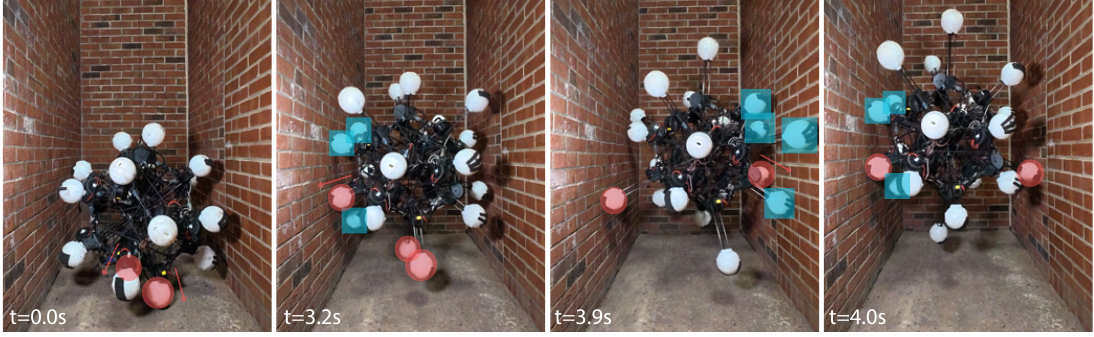}
    \caption{ \textbf{Foot coordination during parallel wall climbing.} Argus initiates climbing by extending its lower foot nodes to anchor between parallel walls. Owing to its near-extreme dynamic isotropy, some feet orient diagonally toward the wall (red), generating linear thrust that produces both upward motion and lateral force pressing the robot against the opposite wall. In turn, other feet (blue) reconfigure into flat contact patches to stabilize the robot. By alternating between thrusting and bracing, Argus ascends steadily along the vertical direction.}
    \label{fig:extended_wall_climbing}
\end{extfigure}

\newpage
\begin{extfigure}[!ht]
    \centering
    \includegraphics[width=1\textwidth]{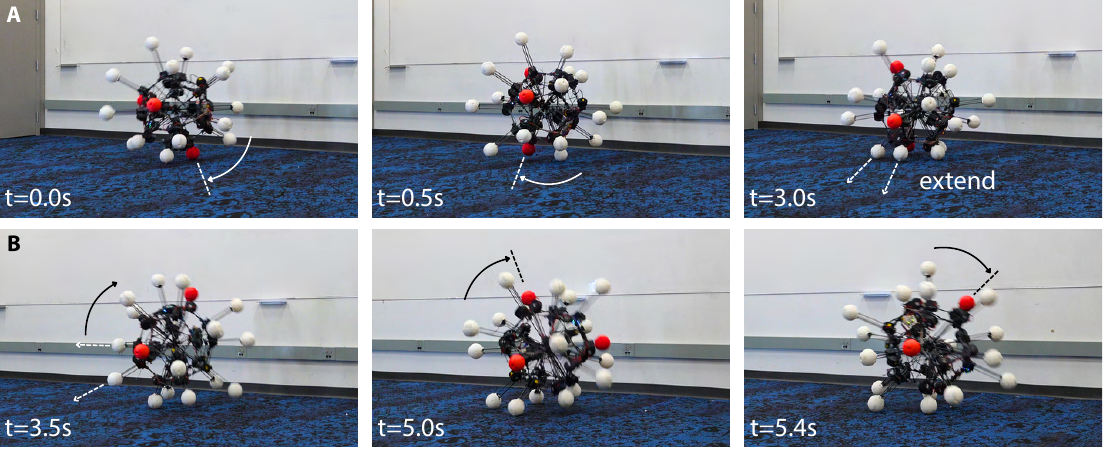}
    \caption{ \textbf{Resilient gaits under actuator failure.} Benefit from the uniformly arranged redundant actuations, Argus can redistribute actuation loads to compensate for the disabled legs. \textbf{(A)}, With one leg disabled at the bottom, Argus engaged the adjacent two legs to restore its rolling gait. \textbf{(B)}, When three legs are disabled, Argus oriented its body to avoid the usage of these three legs.}
    \label{fig:extended_disabled_legs}
\end{extfigure}

\newpage
\begin{extfigure}[!ht]
    \centering
    \includegraphics[width=1\textwidth]{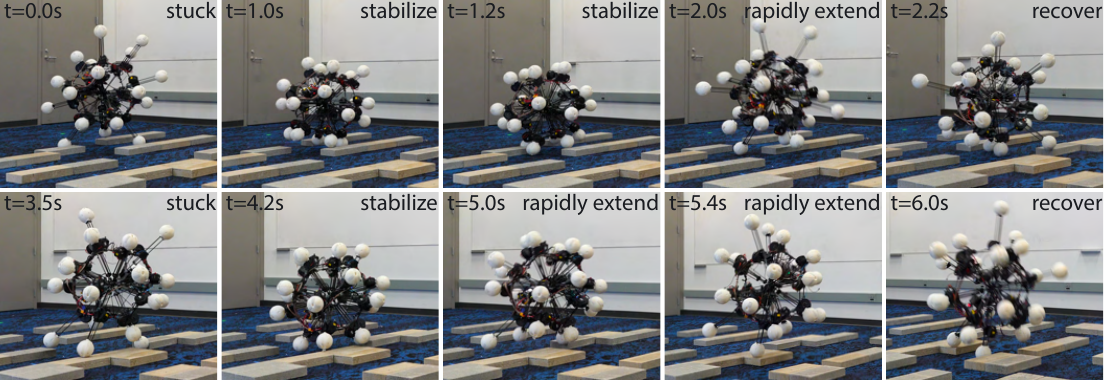}
    \caption{ \textbf{Robust gaits on discrete terrain.} Leveraging its isotropic leg distribution, Argus recovered when stuck on irregular blocks by retracting into a stable posture first, and then rapidly extending its legs to push itself over the obstacle. This thrust generated momentum that propels the body forward.}
    \label{fig:extend_discrete_terrain}
\end{extfigure}

\newpage
\begin{extfigure}[!ht]
    \centering
     \includegraphics[width=1\textwidth]{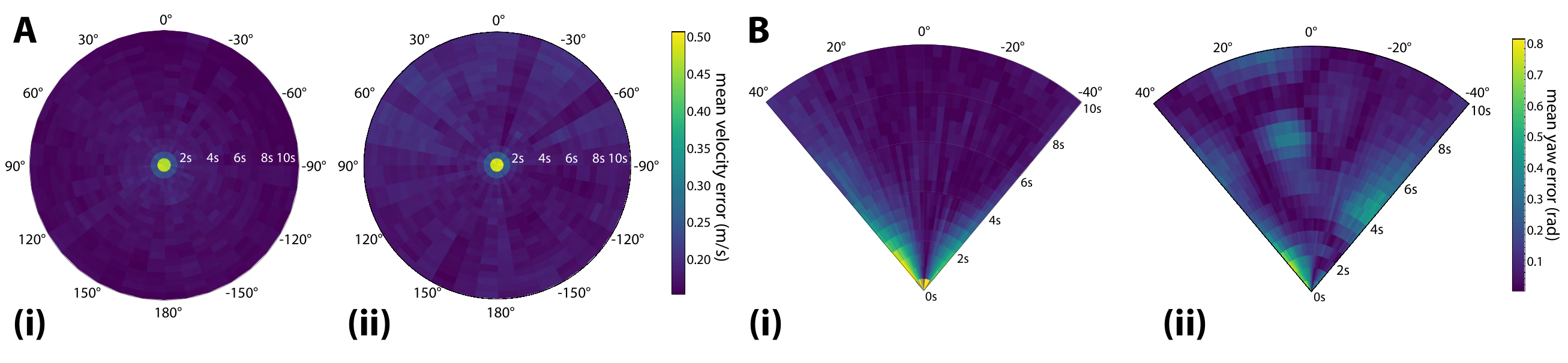}
    \caption{ \textbf{Object tracking and pushing performance.} \textbf{(A)}, Mean velocity error across ten seconds for all sampled commanded directions, shown for the base policy (i) and the final policy with a trained point cloud encoder (ii). Commanded velocities were discretized along both angular and radial dimensions into 32 angular and 20 temporal bins on a polar map. Errors were initially high as the robot accelerated to match the commanded velocity, but rapidly dropped within two seconds. The point cloud-based policy achieved performance comparable to the base policy, with an error of \SI{0.167 \pm 0.120}{\meter/\second}, compared to the base policy, which has an error of \SI{0.162 \pm 0.116}{\meter/\second}. \textbf{(B)}, Mean yaw error of object states for commanded pushing directions sampled from \SI{-40}{\degree},\SI{40}{\degree}] with the same binning approach as above. Error increased near the extremes due to greater efforts to reorient the object. The policy using point cloud observation achieved a mean tracking error of \SI{0.256 \pm 0.207}{\radian}, which is similar to the base policy’s \SI{0.268 \pm 0.227}{\radian}. However, it exhibits less consistent pushing behavior, as indicated by the more irregular color patterns in the heatmap. This is expected due to the reliance on pure point cloud observations without privileged information of object states.}
  \label{fig:extend_object_tracking_eval}
\end{extfigure}

\newpage

\begin{extfigure}[!ht]
    \centering
    \includegraphics[width=1\textwidth]{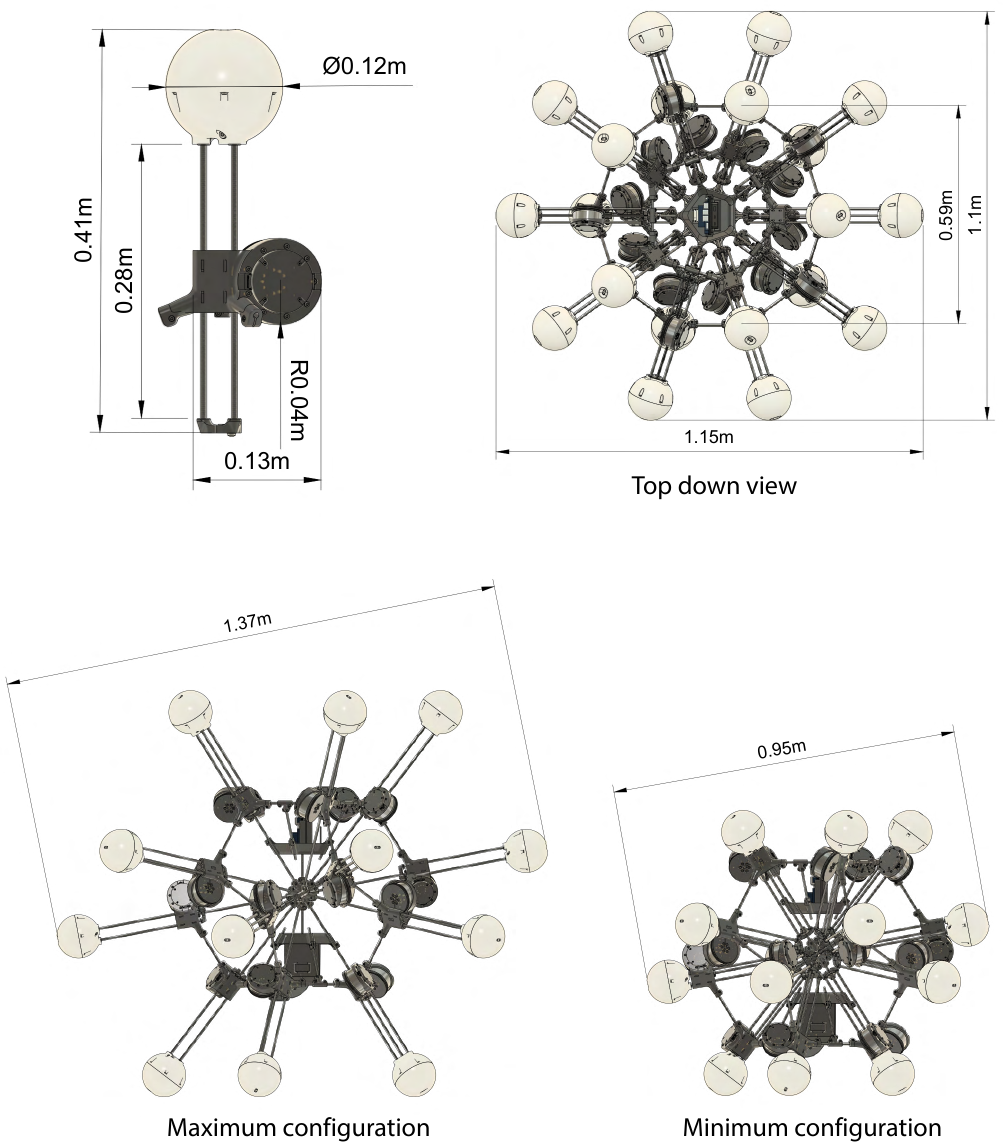}
    \caption{ \textbf{Mechanical drawing of the Argus actuator and assembly with dimensions.}}
    \label{fig:extended_argus_sizing}
\end{extfigure}

\clearpage

\begin{exttable}[!ht]
\centering
\caption{\textbf{Observations and states.}}
\resizebox{1\textwidth}{!}{%
\begin{tabular}{>{\raggedright\arraybackslash}p{3cm} >{\raggedright\arraybackslash}p{7cm} >{\raggedright\arraybackslash}p{7cm}}
\toprule
\textbf{Task} & \textbf{Observations} & \textbf{States} \\
\midrule
all rolling locomotion & 
three-frame proproceptive history &
\textbf{observation}, linear velocity, projected gravity, foot contact, velocity command \\
\midrule

object pushing(stage-1) &
proproceptive, object goal velocity, object orientation &
\textbf{observation}, linear velocity, projected gravity, foot contact, object velocity, robot position, object position \\

object tracking(stage-1) &
proproceptive, object velocity &
\textbf{observation}, linear velocity, projected gravity, object velocity \\

object pushing(stage-2) &
proproceptive, object goal velocity, history of point clouds &
\textbf{observation}, linear velocity, projected gravity, foot contact, object velocity, robot position, object position \\

object tracking(stage-2) &
proproceptive, three-frame point cloud history &
\textbf{observation}, linear velocity, projected gravity, object velocity \\

\bottomrule
\end{tabular}
}
\label{table:obs_state}
\vspace*{-15pt}
\end{exttable}

\clearpage

\begin{exttable}[!ht]
\centering
\caption{\textbf{Domain randomization parameters.}}
\begin{tabular}{llll}
\toprule
Parameter & Unit   & Range & Operator \\
\midrule
Friction            & -             & $[0.1, 0.8]$      & -                 \\
Base COM            & \si{\meter}              & $[-0.05,0.05]$     & additive         \\
Base mass           & \si{\kilogram}             & $[-0.5, 0.5]$        & additive         \\
Link mass           & \si{\kilogram}             & $[0.95, 1.05]$    & scaling          \\
Link inertia        & \si[inter-unit-product = \ensuremath{\cdot}]{\kilogram\meter\squared}  & $[0.95, 1.05]$    & scaling          \\
Joint offset        & \si{\meter}           & $[-0.01, 0.01]$   & additive          \\
Joint position      & \si{\meter}           & $[-0.005, 0.005]$   & additive          \\
Joint velocity      & \si{\meter/\second}         & $[-0.2, 0.2]$     & additive          \\
Motor strength      & -             &  $[0.98, 1.02]$       & scaling       \\
Observation delay   & \si{\milli\second}            & $[0, 40]$         & -                 \\
Angular velocity    & \si{\radian/\second}         & $[-0.2, 0.2]$     & additive          \\
% Gravity vector      & -             & $[-0.05, 0.05]$   & additive          \\
PD Factors          & -            & [0.9, 1.1]       & scaling           \\ 
\bottomrule
\end{tabular}
\label{table:randomization}
\vspace{-5pt}
\end{exttable}

\clearpage

\begin{exttable}[!ht]
% \vspace*{-15pt}
\centering
\caption{\textbf{Baseline RL policy reward functions.}}
\resizebox{0.7\textwidth}{!}{%
\begin{tabular}{rcr}
\toprule
 Reward Terms & Definition & Weight \\
\midrule
Linear velocity & $\phi\left(\mathbf{v}_b^*-\mathbf{v}_b,-5*[1,1,0.1] \right)$& $1 $ \\
% Angular velocity  & $\phi\left(\mathbf{\omega}_b^*-\mathbf{\omega}_b,-8*[0.1,0.1,1] \right)$& $0.5 $ \\
% Base height  & $\phi\left(h_b^*-h_b,-2000 \right)$& $0.1$\\
%  Orientation & $max \left( \left\|\mathbf{g}_{b,xy}\right\|^2, 0.1\right)$&$-20$\\
% \midrule
% Joint power  & $\|\dot{\mathbf{q}}\cdot\mathbf{ \uptau}\|_1$& $-0.0005$\\
Joint velocity & $\phi\left(\mathbf{\dot{\boldsymbol{\mathrm{q}}}},-0.2 \right)$ & $0.1$ \\
% Joint acceleration  & $\phi \left( \ddot{\mathbf{q}}, -10^{-4} \right)$ & $0.1 $ \\
Joint torque  & $ \phi \left(  \uptau - \text{CLIP}\left( \uptau/\uptau_{\max},-0.3,0.3  \right),  -0.1 \right)$ & $0.4$\\
Joint limit & $ \left\|  \mathbf{q}  - \text{CLIP}\left( \mathbf{q},\mathbf{q}_{\min},\mathbf{q}_{\max}  \right)  \right\|^2$ & $-1$ \\
Action rate  & $\phi\left(\mathbf{\dot{a}},-0.0002 \right)$ & $0.2$\\
Action & $ \phi \left(  \mathbf{a} - \text{CLIP}\left(\mathbf{a},-0.5,0.5  \right),  -0.2 \right)$ & $0.2$ \\
Foot impact & $\left\|\text{CLIP}(\textbf{F}_{cz}/G -1, 0, 2)\right\|^2$ & $-0.01$ \\
Foot slip & $\left\| \mathbf{v}_{f}\cdot\mathbf{I}\right\|^2$ & $-0.01$ \\
\midrule
Object orientation  & $\text{CLIP}( -1/|{\mathbf{y}_{obj}} \cdot 
{\mathbf{v}_{command}}|+1,-0.2,0)$ & $2$\\
Robot-object distance  & $\phi\left(\mathbf{p}_{robot}-\mathbf{p}_{object},-2 \right)$ & $0.8$\\

\bottomrule
\end{tabular}
}
\label{table:baseline_reward}
\vspace*{-15pt}
\end{exttable}

\newpage

\begin{exttable}[!ht]
\centering
\caption{\textbf{Comparison of spherical and tensegrity robotic actuators.}}
\resizebox{1.0\textwidth}{!}{%
\begin{tabular}{p{3.5cm} c p{4.5cm} c c c c c}
\toprule
\makecell{\textbf{Robot} \\ \textbf{Name}} & 
\makecell{\textbf{Number of} \\ \textbf{Actuators}} & 
\makecell{\textbf{Actuator Type}} & 
\makecell{\textbf{Stroke} \\ \textbf{(m)}} & 
\makecell{\textbf{Max} \\ \textbf{Thrust (N)}} & 
\makecell{\textbf{Actuator} \\ \textbf{Weight (kg)}} & 
\makecell{\textbf{Max} \\ \textbf{Velocity (m/s)}} & 
\makecell{\textbf{Max Thrust per} \\ \textbf{Actuator Mass (N/kg)}} \\
\midrule
Argus (This work*) & 20 & cable driven & 0.21 & 375 & 0.62 & \textbf{1.0} & \textbf{604.8} \\
Six-strut \cite{zheng2021robustness} & 6 & linear actuator & 0.10 & 15 & 0.065 & 0.014 & 230.8 \\
Spiny \cite{nozaki2017Shape} & 12 & telescopic linear actuator & 0.32 & 25 & 0.4 & 0.2 & 62.5 \\
Mochibot \cite{nozaki2018Continuous} & 32 & telescopic linear actuator & 0.32 & 25 & 0.4 & 0.2 & 62.5 \\
SUPERball v2 \cite{vespignani2018design} & 24 & cable driven & 0.97 & 375 & 1.5 & 0.14 & 250.0 \\
Multiretractable \cite{xu2024physics} & 12 & telescopic linear actuator & 0.20 & 50 & 0.15 & 0.006 & 333.3 \\
Sea urchin \cite{yang2023bionic} & 12 & linear actuator & 0.07 & 35 & 0.75 & 0.1 & 46.7 \\
\bottomrule
\end{tabular}
}
\label{table:robot_actuators}
\vspace*{-15pt}
\end{exttable}

\newpage
\subsection*{Captions for Movies}

\mypara{Movie 1 | Overview of robotic system.}
The overview of the paper.

\mypara{Movie S1 | Flat Terrain Velocity Tracking}
Argus follows square and arbitrary trajectories on flat terrain to demonstrate velocity control.

\mypara{Movie S2 | Locomotion with leg failure}
Argus walks with one, two, and three legs randomly disabled, each from a different starting orientation.

\mypara{Movie S3 | Self-Stabilization}
Argus stabilizes itself against disturbances, comparing performance with and without active control.

\mypara{Movie S4 | Carrying Objects}
Argus walks while carrying a loaded pouch with weights of 2, 4, 6, 8, and 10 pounds.

\mypara{Movie S5 | Discrete Terrain Traversal}
Argus traverses discrete obstacle terrains from varying initial orientations.

\mypara{Movie S6 | Outdoor Terrain -- Duke Campus}
Argus navigates concrete, grass, and dirt surfaces on the Duke University campus, filmed in both stationary and tracking views.

\mypara{Movie S7 | Outdoor Terrain -- Sand and Forest}
Argus walks across sand and forest trails, filmed from a tracking perspective.

\mypara{Movie S8 | Wall Climbing under Moon Gravity}
Argus climbs a vertical wall using a gravity-assist pulley simulating lunar gravity.

\mypara{Movie S9 | Autonomous Object Tracking}
Argus tracks a cube programmed to move in constant and random directions using onboard cameras.

\mypara{Movie S10 | Autonomous Object Pushing}
Argus pushes a cube in three directions: 0$^\circ$, -30$^\circ$, and +30$^\circ$ using onboard cameras.

%%%%%%%%%%%%%%%% SUPPLEMENTARY REFERENCES %%%%%%%%%%%%%%%

% Do NOT include a reference list in the supplement.
% All references must be in a single list at the end of the main text.
% The copyeditors will ensure that the correct reference list appears with each version of the paper
% (print, HTML, PDF, mobile app, metadata for bibliographic databases etc.)

\end{document}